\pdfoutput=1

\documentclass[10pt,twocolumn,letterpaper]{article}

\usepackage[pagenumbers]{wacv} 

\usepackage{background}

\backgroundsetup{angle=0, 
scale=1,
color=black,
firstpage=true,
position=current page.north,
hshift=0pt,
vshift=-20pt,
contents={\ifnum\value{page}=1 PREPRINT: Accepted at the 2024 IEEE/CVF Winter Conference on Applications of Computer Vision (WACV 2024) \else \fi}
}

\usepackage{times}  
\usepackage{helvet}  
\usepackage{courier}  
\usepackage[hyphens]{url} 
\usepackage{graphicx} 
\usepackage[numbers]{natbib}  
\usepackage{caption} 
\usepackage{amsmath}
\usepackage{amssymb}
\usepackage{booktabs}
\usepackage{amsfonts}
\usepackage{multirow}
\usepackage{arydshln}
\usepackage{soul}
\usepackage{xcolor}
\usepackage[ruled,linesnumbered]{algorithm2e}
\SetKwComment{Comment}{$\triangleright$\ }{}
\usepackage{algorithmicx}
\usepackage{algpseudocode}
\usepackage{nicefrac}
\usepackage{subcaption}
\usepackage[accsupp]{axessibility}

\newlength\mylenin
\newcommand\myinput[1]{%
\settowidth\mylenin{\KwIn{}}%
\setlength\hangindent{\mylenin}%
\hspace*{\mylenin}#1\\}

\let\oldnl\nl 
\newcommand{\nonl}{\renewcommand{\nl}{\let\nl\oldnl}} 

\newlength\mylenout

\DeclareMathOperator*{\argmin}{arg\,min}
\DeclareMathOperator{\E}{\mathbb{E}}

\usepackage[pagebackref,breaklinks,colorlinks]{hyperref}

\usepackage[capitalize]{cleveref}
\crefname{section}{Sec.}{Secs.}
\Crefname{section}{Section}{Sections}
\Crefname{table}{Table}{Tables}
\crefname{table}{Tab.}{Tabs.}

\def\wacvPaperID{459} 
\def\confName{WACV}
\def\confYear{2024}

\begin{document}

\title{Meta-Learned Kernel For Blind Super-Resolution Kernel Estimation}

\author{%
  \textbf{Royson Lee\textsuperscript{1,2}\thanks{Corresponding Author: \texttt{dsrl2@cam.ac.uk}} , Rui Li\textsuperscript{2}, Stylianos Venieris\textsuperscript{2}} \\
  \textbf{Timothy Hospedales\textsuperscript{2,3}, Ferenc Huszár\textsuperscript{1}, Nicholas D. Lane\textsuperscript{1,4}} \\
  \\
  \textsuperscript{1} University of Cambridge, UK \hspace*{10pt}
  \textsuperscript{2} Samsung AI Center, Cambridge, UK\\
  \textsuperscript{3} University of Edinburgh, UK \hspace*{10pt}
  \textsuperscript{4} Flower Labs \\
}

\maketitle

\begin{abstract}
    Recent image degradation estimation methods have enabled single-image super-resolution (SR) approaches to better upsample real-world images. Among these methods, explicit kernel estimation approaches have demonstrated unprecedented performance at handling unknown degradations. Nonetheless, a number of limitations constrain their efficacy when used by downstream SR models. Specifically, this family of methods yields \textit{i)}~excessive inference time due to long per-image adaptation times and \textit{ii)}~inferior image fidelity due to kernel mismatch. In this work, we introduce a learning-to-learn approach that meta-learns from the information contained in a distribution of images, thereby enabling significantly faster adaptation to new images with substantially improved performance in both kernel estimation and image fidelity. Specifically, we meta-train a kernel-generating GAN, named MetaKernelGAN, on a range of tasks, such that when a new image is presented, the generator starts from an informed kernel estimate and the discriminator starts with a strong capability to distinguish between patch distributions. Compared with state-of-the-art methods, our experiments show that MetaKernelGAN better estimates the magnitude and covariance of the kernel, leading to state-of-the-art blind SR results within a similar computational regime when combined with a non-blind SR model. Through supervised learning of an unsupervised learner, our method maintains the generalizability of the unsupervised learner, improves the optimization stability of kernel estimation, and hence image adaptation, and leads to a faster inference with a speedup between $14.24$ to $102.1\times$ over existing methods.\footnote[0]{Code is available at https://github.com/royson/metakernelgan}
\end{abstract}
\vspace{-2em}
\section{Introduction}\label{sec:introduction}


Single-image super-resolution (SR) is a low-level vision task that entails the upsampling of a low-resolution (LR) image to high resolution (HR). Despite the significant progress of deep learning-based SR in recent years, the majority of existing approaches~\cite{SRCNN,FSRCNN,ESPCN,TPSR,ESRN,CARN,Lee2019,Lee2021DeepNN} assume a \textit{fixed} degradation process with a known blur kernel, such as bicubic interpolation. This fact often leads to poor performance in real-world cases, due to the variability of the degradation process across images, and puts a strong \textit{retraining} requirement for every new degradation operation.

In this context, an alternative line of SR methods has emerged that aims to improve the performance under unknown degradations. These methods can be taxonomized into non-blind and blind.
Non-blind SR leverages a \textit{provided} degradation process in order to maximize the upsampling quality for the given image.
Blind SR, on the other hand, \textit{estimates} the degradation process.

\begin{figure}
    \centering
    {
    \includegraphics[width=0.4\textwidth]{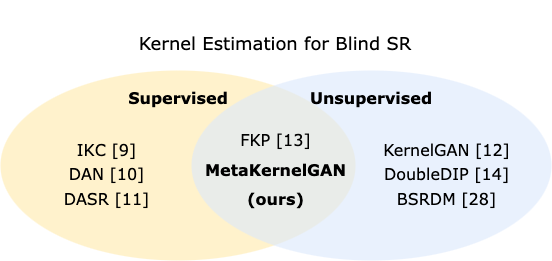}
    }
    \caption{
    Landscape of kernel estimation methods for blind SR. Supervised methods are computationally efficient, but underperform when faced with out-of-distribution kernels on real-world images. Unsupervised methods can better adapt to diverse kernels, but are undeployable due to their excessive computational overhead. Our MetaKernelGAN method blends together both approaches for the first time and introduces the essential techniques to obtain the respective benefits.
    }
    \label{fig:blindsr}
    \vspace{-2em}
\end{figure} 

Estimating the degradation process can be done either implicitly or explicitly. 
Implicit methods typically adopt a \textit{supervised learning} approach where the degradation representation is learnt jointly with an SR model on a given dataset~\cite{IKC,DAN,Wang2021UnsupervisedDR} (Fig.~\ref{fig:blindsr} - left). In such schemes, the features of the degradation process of the given training set are implicitly built into the SR model, yielding high performance on test images with the same degradation. From a computational aspect, as no adaptation is performed at inference time, these methods only require a forward pass and hence are fast. Nonetheless, the main limitation of these methods is that their performance deteriorates significantly when the test degradation differs from the train distribution. This property makes them unsuitable for the variably degraded images in the wild and renders their applicability narrow to settings with well-known degradation processes.

In contrast, explicit degradation estimation methods~\cite{KernelGAN,FKP} learn a degradation process\footnote{Generally the blur kernel is assumed to be anisotropic Gaussian.} for each given image at inference time. These methods either jointly optimize the super-resolved image together with the kernel or feed both the kernel and the given image to a downstream non-blind SR model in a two-step process. The degradation estimation is performed through \textit{unsupervised learning} and requires a number of training iterations to effectively learn the image-specific degradation process (Fig.~\ref{fig:blindsr} - right).
By adapting on each image, these methods not only lead to better performance on unseen degradation distributions, but also provide a better interpretation of the kernel in the pixel space through their explicit degradation modeling.
Nonetheless, the additional computations per image and their unsupervised nature make these methods slow and unstable, often requiring thousands of training steps for each image to achieve adequate performance.

To counteract the limitations of these unsupervised methods, existing approaches constrain the set of possible kernels solutions by either pretraining the model to start with an initial Gaussian-like kernel~\cite{FKP}, jointly optimizing both the image and kernel by minimizing the LR image reconstruction loss~\cite{Gandelsman2019DoubleDIPUI}, explicitly optimizing the precision matrix of the Gaussian kernel~\cite{BSRDM}, or heavily enforcing the use of regularization, such as encouraging the kernel to be a bicubic kernel at the start~\cite{KernelGAN}. Although effective, the aforementioned limitations are not sufficiently mitigated, especially in cases where the given image does not contain sufficient information, \textit{e.g.} noisy and/or tiny images.
 
In this work, we aim to tackle the same set of challenges of unsupervised kernel estimation through supervised learning while retaining the benefits of unsupervised learning.
In other words, our method should only utilize the internal information of the given image during inference to better handle unseen degradations but fall back on a learned kernel that works well on average across a dataset of images in cases where this internal information is limited (see Fig.~\ref{fig:blindsr} - center).
To this end, we propose MetaKernelGAN, a gradient-based meta-learning approach that learns from the unsupervised kernel estimation learner. Concretely, we build upon KernelGAN~\cite{KernelGAN} as our unsupervised learner and aim to meta-learn the initialization of both the generator and discriminator so that it can rapidly and accurately estimate the kernel for each image. 
Through extensive experiments, we demonstrate the benefits, limitations, and suitability of our approach, compared with existing kernel estimation and blind SR approaches on both standard datasets and real-world images. 
We also provide a detailed analysis of the kernel estimation process, highlighting the advantages and limitations of meta-learning in explicit kernel estimation.
Most importantly, we show that this blending of supervised and unsupervised learning holds considerable potential in utilizing external images for unsupervised degradation estimation works, directly counteracting the fundamental limitations of these per-image methods.

\section{Related Work}\label{sec:related_work}


\noindent
\textbf{Explicit Kernel Estimation.}~Recent kernel estimation approaches have been dominated by the use of deep learning. 
KernelGAN~\cite{KernelGAN} exploits the internal patch recurrence of the image and derives a kernel that preserves the distribution of the image across scales.
Double-DIP~\cite{Gandelsman2019DoubleDIPUI,Ren2020NeuralBD} jointly optimizes the HR image and the blur kernel by minimizing the LR image reconstruction loss. 
These explicit kernel estimation works learn the kernel in the pixel space using neural networks, starting from a random initialization for each image. Hence, they often require multiple regularization terms to restrict the possible set of solutions, resulting in unstable and slow convergence. To counteract this, FKP~\cite{FKP} constrains the kernel space to an anisotropic Gaussian distribution by mapping a simpler latent distribution to a Gaussian distribution through normalizing flows, which is pretrained in a supervised manner. 
Optimizing directly on this latent space traverses a learned kernel manifold that improves accuracy when applied to existing kernel estimation methods, such as KernelGAN and Double-DIP. 
Lastly, BSRDM~\cite{BSRDM} explicitly learns the precision matrix of the Gaussian kernel and models the noise variance.
Instead of learning a restricted space of Gaussian kernel solutions, our approach meta-learns ``how to adapt" across multiple unsupervised adaptation steps using a diverse distribution of images.
\noindent
\textbf{Meta-Learning.}~Meta-learning methods aim to extract transferable knowledge from a distribution of tasks so as to solve new tasks more data efficiently. This area is now too wide to review exhaustively here, and we refer the reader to the survey~\cite{hospedales20201metaSurveyPAMI}. Gradient-based meta-learners such as MAML~\cite{MAML} are a popular approach that aim to learn a neural network initialization that can be well adapted with a small amount of data and few gradient steps. In the context of zero-shot super-resolution (ZSSR~\cite{ZSSR}), a few studies such as MZSR~\cite{Soh2020} and MLSR~\cite{Park2020} have applied such meta-learning to learn initializations of SR networks that can be rapidly adapted, accelerating inference compared to prior ZSSR. However, these works are non-blind; they assume the degradation process is known and hence complementary to our work. With respect to GANs, a few studies~\cite{cloutre2019figrMetaGAN,liang2020dawson} have tried with limited success to meta-learn GAN initalizations for data-efficient adaptation of generative models to new categories or domains. In this study, we focus on kernel estimation for blind SR, and for the first time use meta-learning to both accelerate and improve the accuracy of kernel estimation, and hence downstream non-blind super-resolution. Uniquely, we meta-learn the initialization of a kernel-generating GAN from an external distribution of images, so that it can be easily adapted to each testing image with the limited information available from its internal patch recurrence.

\begin{figure}
    \centering
    {
    \includegraphics[trim={0cm 1.5cm 0cm 0cm}, clip, width=0.49\textwidth]{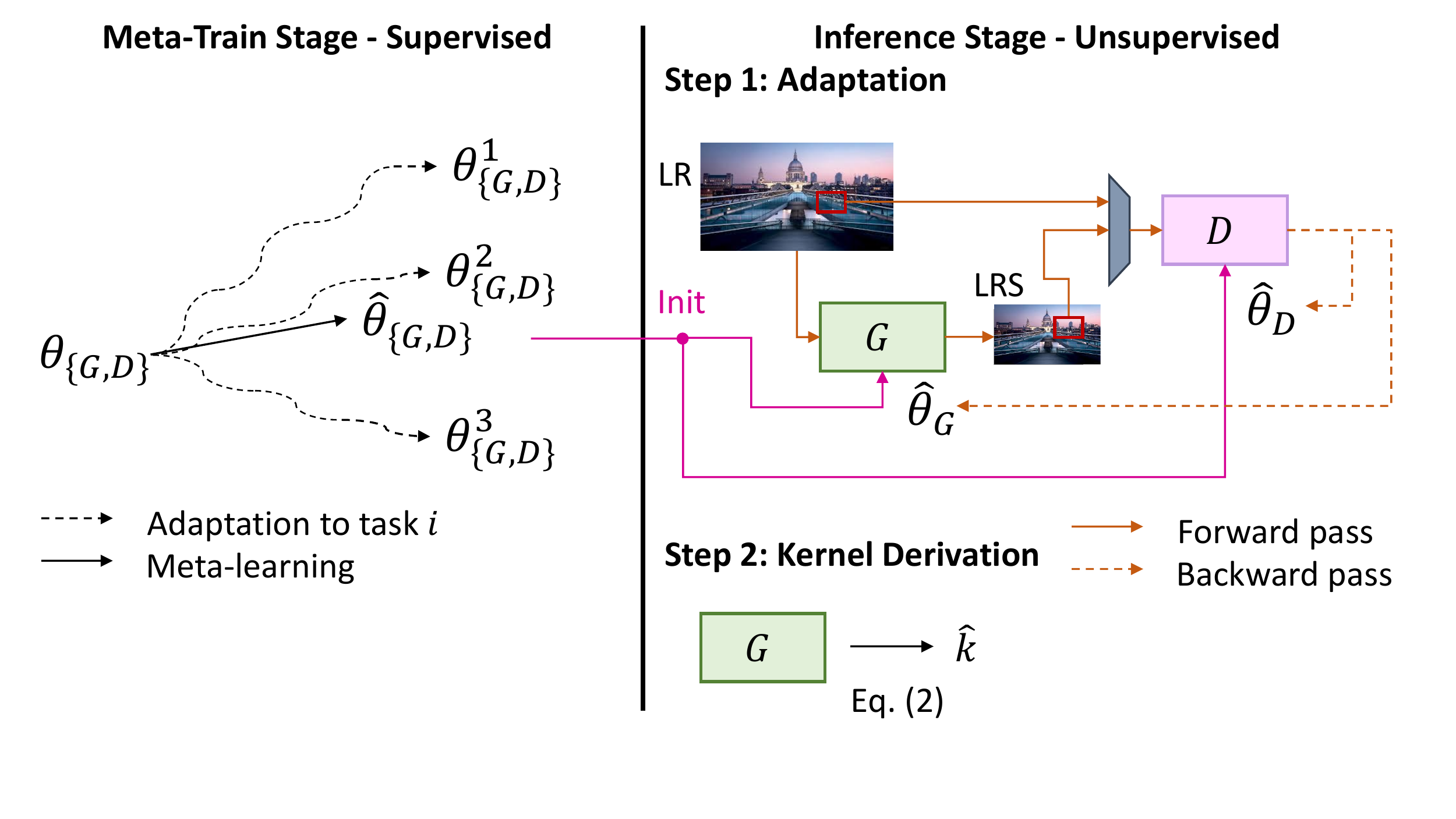}
	\vspace{-4mm}
    }
    \caption{Overview of MetaKernelGAN. }
    \label{fig:metakernelgan}
    \vspace{-1em}
\end{figure}

\section{Meta-Learning for Kernel Estimation}\label{sec:approach}

\noindent
\textbf{Preliminaries.}
The degradation process of image super-resolution is commonly expressed as:
\begin{equation}\label{eq:sr}
I^{\text{LR}} = \big(I^{\text{HR}} * k\big)\downarrow_s +~n
\end{equation}
where $I^{\text{LR}}$ and $I^{\text{HR}}$ are the LR and HR image respectively, $k$ is the blur kernel, $\downarrow_s$ is the subsampling operation with scaling factor $s$, and $n$ is the additive noise. Blind SR aims to estimate both the blur kernel and HR image and the objective is formulated as follows:
\begin{equation}\label{eq:opt_sisr}
    \small
    \hat{I}^{\text{HR}}, \hat{k} = \argmin_{I^{\text{HR}}, k} ||\big(I^{\text{HR}} * k\big)\downarrow_s - I^{\text{LR}}||_p + \phi(I^{\text{HR}}) + \Phi(k)
\end{equation}
where $||\big(I^{\text{HR}} * k\big)\downarrow_s - I^{\text{LR}}||_p$ is the fidelity term, $\phi(I^{\text{HR}})$ is the image prior, and $\Phi(k)$ is the kernel prior. Image priors have been well-investigated and are often implicitly represented through convolutional neural networks (CNNs) as they \textit{1)}~constitute structurally strong natural image priors~\cite{deep_image_prior} and \textit{2)}~are highly effective at learning priors from a large set of LR-HR pairs~\cite{SRCNN}.
In contrast, kernel priors have only been recently studied through the use of fully-connected layers~\cite{Ren2020NeuralBD}, a deep linear convolutional network~\cite{KernelGAN}, or normalizing flows~\cite{FKP}, all of which solely rely on the internal information within a given image. In this work, we focus on estimating the kernel and use an existing non-blind SR work for upsampling.

\noindent
\textbf{KernelGAN for Internal Learning.}~A prominent approach to estimate the kernel from an LR image is by exploiting its internal patch recurrence. Specifically, KernelGAN~\cite{KernelGAN} achieves this by employing a deep linear generator as a prior to learn the linear degradation process (Eq.~(\ref{eq:sr})). First, the generator, $G$, takes in a patch from the given LR image and downsamples it to obtain the resulting LR son (LRS) patch. Random patches are sampled from the LR image, \textit{i.e.}~$p^{\text{LR}}$$\sim$$patches(I^{\text{LR}})$, and the corresponding LRS patches are generated by the generator $G$, \textit{i.e.}~$G(p^{\text{LR}};\theta_G)$. These are then fed into the discriminator, $D$, which distinguishes between the two patch distributions. 
After training, the estimated kernel is derived from the generator as follows:

\begin{equation}\label{eq:kernel_deriv}
    \small
    \hat{k} = \text{DK}(\theta_G) =  \hat{G}(\mathbf{J}_{\lceil\frac{m}{2}\rceil\lceil\frac{m}{2}\rceil};\theta_G)
\end{equation}
where $\hat{k}$$\in$$[0,1]^{m \times m}$ is the estimated kernel, $\hat{G}$ is $G$ with its convolutional strides set to $1$, and {\small$\mathbf{J}_{\lceil\frac{m}{2}\rceil\lceil\frac{m}{2}\rceil}$$\in$$\{0,1\}^{(2m-1)\times (2m-1)}$} is a single-entry matrix with all zeros except for a value of 1 at its central element. 

As KernelGAN directly estimates the kernel in the pixel space, it employs multiple regularization terms to restrict the number of possible solutions to a Gaussian-like kernel such as being similar to the bicubic kernel at the start, incentivizing the optimal estimated kernel, $k^*$, to sum to one, discouraging boundary pixels, and encouraging sparsity and centrality.
The need to retrain both the generator and discriminator from scratch and tune the hyperparameters of these regularization terms per image constitutes two of the main limitations of KernelGAN, resulting in unstable results and excessively long inference times.

\setlength{\textfloatsep}{0pt}
\SetArgSty{textnormal}
\begin{algorithm}[!t]
    \scriptsize
    \SetAlgoLined
    \LinesNumbered
    \DontPrintSemicolon
    \KwIn{Distribution over tasks $p(\mathcal{T})$}
    \nonl
    \myinput{Number of steps $N_{\text{steps}}, N_{\text{adapt}}, N_{\text{val}}$}
    \nonl
    \myinput{Step size hyperparameters $\alpha_G, \alpha_D, \beta_G, \beta_D$}
    \nonl
    \myinput{Loss weighting hyperparameters $\omega, \eta, \zeta$}
    \KwOut{Meta-learned $\hat{\theta}_G$ and $\hat{\theta}_D$}
    
    Initialize $\theta_G$ and $\theta_D$ \\
    \For(\Comment{\scriptsize \textrm{Meta-optimization steps}}){$j$ in $[1,N_{\text{steps}}]$}{
        $\theta^i_G \leftarrow \theta_G$, \quad $\theta^i_D \leftarrow \theta_D$ \\
        Sample a batch of tasks $\mathcal{T}^i = \left< I^{\text{LR},i}, k \right> \sim p(\mathcal{T})$ \\
        Interval loss dictionary $b \leftarrow \{ \cdot \}$ \\
        \ForAll{$\mathcal{T}^i$}{
            $b[\mathcal{T}^i][G] \leftarrow [\cdot]$, \quad $b[\mathcal{T}^i][D] \leftarrow [\cdot]$  \\
            \For(\Comment{Adaptation steps over task $i$}){$l$ in $[1,N_{\text{adapt}}]$}{
                Compute adapted parameters 
                $\left(\text{Eq.}~(\ref{eq:eq_inner_loop})\right)$: \newline {$\theta_G^i \leftarrow \theta_G^i - \alpha_G \nabla_{\theta_G} \mathcal{L}_G^{\text{task}}(\theta_D^i, \theta_G^i)$} 
                \newline
                {$\theta_D^i \leftarrow \theta_D^i - \alpha_D \nabla_{\theta_D} \mathcal{L}_D^{\text{task}}(\theta_D^i, \theta_G^i)$}
                
                \If(\Comment{Interval loss evaluation}){$(l\ \text{mod}\ N_{\text{val}}) = 0$}{
                    $b[\mathcal{T}^i][G]\text{.append}\left( \mathcal{L}_G^{\text{meta}}(\theta_D^i, \theta_G^i, k) \right)$ 
                    \\
                    $b[\mathcal{T}^i][D]\text{.append}\left( \mathcal{L}_D^{\text{meta}}(\theta_D^i, \theta_G^i) \right)$
                }
                
                %
                %
            }
        }
        $\mathbf{w} \leftarrow \text{GetIntervalLossWeights}(j)$ \\
        Update $\theta_G$ and $\theta_D$: \newline
        {$
            \theta_G \leftarrow \theta_G - \beta_G \nabla_{\theta_G} \sum_{\mathcal{T}^i \sim p(\mathcal{T})} \big( b[\mathcal{T}^i][G] \odot \mathbf{w} \big)
        $} \newline 
        $\theta_D \leftarrow \theta_D - \beta_D \nabla_{\theta_D} \sum_{\mathcal{T}^i \sim p(\mathcal{T})} \big( b[\mathcal{T}^i][D] \odot \mathbf{w} \big)$
    }
    $\hat{\theta}_G \leftarrow \theta_G$,
    $\hat{\theta}_D \leftarrow \theta_D$
    \caption{Meta-Training of MetaKernelGAN}
    \label{alg:mlkp_train}
\end{algorithm}



\begin{table}[t]
\centering
\resizebox{0.48\textwidth}{!}{%
\begin{tabular}{l|l|l|l|l|l} \toprule
\multicolumn{1}{c}{\textbf{Method}} & \multicolumn{1}{c}{\textbf{Steps}} & \multicolumn{1}{c}{\textbf{Set14}} & \multicolumn{1}{c}{\textbf{B100}} & \multicolumn{1}{c}{\textbf{Urban100}} & \multicolumn{1}{c}{\textbf{DIV2K}} \\ \midrule
\multicolumn{6}{c}{$\times$2} \\ \hline
Bicubic & - & 24.82/0.6910/-/- &25.17/0.6603/-/- &22.31/0.6471/-/- &26.94/0.7642/-/-\\
KernelGAN~\cite{KernelGAN} + USRNet & 3000 & 25.11/0.7404/43.37/3.22 & 24.81/0.7081/43.95/3.00 & 23.16/0.7259/44.31/3.10 & 27.56/0.8157/44.72/2.82 \\
KernelGAN-FKP~\cite{FKP} + USRNet & 1000 & 22.11/0.6229/45.25/9.70 & 23.48/0.6628/44.41/6.61 & 22.67/0.7099/44.93/4.77 & 28.83/0.8583/47.94/3.09 \\


DSKernelGAN (our baseline) + USRNet & 200 & 28.34/0.8210/42.17/4.15 & 28.06/0.7928/41.26/4.64 & 25.99/0.8046/42.84/3.74 & 31.07/0.8732/43.54/3.53 \\ 
MetaKernelGAN (ours) + USRNet & 200 & \textbf{28.71}/\textbf{0.8331}/\textbf{46.23}/\textbf{2.44} & \textbf{28.95}/\textbf{0.8222}/\textbf{45.94}/\textbf{2.38} & \textbf{26.78}/\textbf{0.8355}/\textbf{47.37}/\textbf{2.16} & \textbf{31.67}/\textbf{0.8944}/\textbf{48.00}/\textbf{2.08} \\


\hdashline
GT + USRNet (upper bound) & - & 32.56/0.8944/-/- & 31.33/0.8771/-/- & 29.96/0.8954/-/- & 34.59/0.9268/-/- \\ 
\hline
\multicolumn{6}{c}{$\times$2 with Non-Gaussian Kernel} \\ \hline
Bicubic & - & 24.80/0.6907/-/- &25.17/0.6609/-/- &22.31/0.6476/-/- &26.93/0.7640/-/- \\
KernelGAN~\cite{KernelGAN} + USRNet & 3000 & 24.29/0.7088/41.75/3.16 & 23.84/0.6780/42.50/2.94 & 22.12/0.6940/42.86/2.96 & 26.52/0.7937/43.13/2.83 \\
KernelGAN-FKP~\cite{FKP} + USRNet & 1000 & 20.91/0.6034/42.35/13.08 & 22.75/0.6447/42.54/6.31 & 21.54/0.6742/42.85/4.77 & 27.40/0.8339/44.75/3.08 \\


DSKernelGAN (our baseline) + USRNet & 200 & 28.00/0.8067/40.87/4.03 & 27.74/0.7805/39.9/4.83 & 25.51/0.7906/41.48/3.68 & 30.74/0.8672/42.0/3.56 \\ 
MetaKernelGAN (ours) + USRNet & 200 & \textbf{28.05}/\textbf{0.8153}/\textbf{43.68}/\textbf{2.36} & \textbf{28.2}/\textbf{0.8031}/\textbf{43.69}/\textbf{2.32} & \textbf{25.84}/\textbf{0.8122}/\textbf{44.72}/\textbf{2.12} & \textbf{30.96}/\textbf{0.8830}/\textbf{44.96}/\textbf{2.08} \\


\hdashline
GT + USRNet (upper bound) & - & 32.47/0.8968/-/- & 31.34/0.8803/-/- & 30.08/0.8999/-/- & 34.59/0.9272/-/- \\ \hline

\multicolumn{6}{c}{$\times$2 with Image Noise of Level 10 (3.92\%)} \\ \hline
Bicubic & - & 24.69/0.6639/-/- &24.98/0.6321/-/- &22.20/0.6162/-/- &26.65/0.7273/-/- \\
KernelGAN~\cite{KernelGAN} + USRNet & 3000       & 27.29/0.7671/43.15/3.29 & 27.10/0.7353/43.69/3.22 & 25.08/0.7679/44.34/3.10 & 29.50/0.8286/44.66/2.91 \\
KernelGAN-FKP~\cite{FKP} + USRNet & 1000          & 23.23/0.7090/\textbf{44.46}/13.40 & 24.87/0.7188/\textbf{43.72}/7.89 & 23.80/0.7437/44.26/5.04 & 28.57/0.8277/\textbf{47.10}/3.68 \\


DSKernelGAN (our baseline) + USRNet & 200 & 27.57/0.7619/41.27/4.57 & 27.10/0.7143/39.44/5.55 & 25.15/0.7507/41.03/4.55 & 29.74/0.8154/41.81/4.28 \\ 
MetaKernelGAN (ours) + USRNet & 200 & \textbf{28.23}/\textbf{0.7773}/43.99/\textbf{2.76} & \textbf{27.75}/\textbf{0.7354}/43.07/\textbf{2.98} & \textbf{25.86}/\textbf{0.7780}/\textbf{44.8}/\textbf{2.6} & \textbf{30.25}/\textbf{0.8311}/44.91/\textbf{2.64} \\


\hdashline
GT + USRNet (upper bound) & - & 29.57/0.7990/-/- & 28.56/0.7607/-/- & 27.17/0.8114/-/- & 31.27/0.8497/-/- \\ \hline
\multicolumn{6}{c}{$\times$4} \\ \hline
Bicubic & - & 21.15/0.5280/-/- &22.09/0.5119/-/- &19.30/0.4761/-/- &23.20/0.6329/-/-\\
KernelGAN~\cite{KernelGAN} + USRNet & 3000       & Not supported & Not supported & 19.62/0.5183/57.05/13.53 & 23.51/0.6480/57.27/12.30 \\
KernelGAN-FKP~\cite{FKP} + USRNet & 4000         & Not supported & Not supported & Not supported & 25.34/0.7221/60.45/11.81 \\


DSKernelGAN (our baseline) + USRNet & 200 & 24.61/0.6725/58.77/12.97 & 24.30/0.6127/56.91/18.14 &  21.99/0.6134/57.34/14.89 & 26.93/0.7448/58.74/14.10 \\ 
MetaKernelGAN (ours) + USRNet & 200 & \textbf{25.46}/\textbf{0.6960}/\textbf{63.13}/\textbf{6.54} & \textbf{24.36}/\textbf{0.6200}/\textbf{59.84}/\textbf{10.84} & \textbf{22.21}/\textbf{0.6375}/\textbf{61.74}/\textbf{9.01} & \textbf{26.99}/\textbf{0.7600}/\textbf{61.78}/\textbf{8.34} \\


\hdashline
GT + USRNet (upper bound) & - & 27.89/0.7498/-/- & 26.92/0.6986/-/- & 24.95/0.7357/-/- & 29.46/0.8069/-/- \\ \hline
\end{tabular}%
}
\caption{Average Image PSNR/Image SSIM/Kernel PSNR/$\mathcal{L}^{\text{K{\text -}COV}}$ of patch recurrence methods on SR benchmarks across five runs.}

\label{tab:kg_table}
\vspace{1mm}
\end{table}

\subsection{MetaKernelGAN}\label{sec:metakernelgan}

Our proposed approach, MetaKernelGAN, aims to counteract the drawbacks of KernelGAN to effectively improve its utility and enable fast and stable kernel estimation. Specifically, we adopt a variant of KernelGAN and meta-train the parameters of \textit{both} $G$ and $D$ using a set of diverse images and kernels (Fig.~\ref{fig:metakernelgan}). During inference, our approach is fully unsupervised and is similar to KernelGAN. Unlike KernelGAN, $G$ is initialized with a meta-learned parameter that can adapt rapidly to an accurate kernel and $D$ is initialized such that it can keep up to prevent the generator from dominating. Hence, MetaKernelGAN does not rely on the heavy use of regularization and require fewer hyperparameters, resulting in a more stable adaptation process and needing significantly fewer adaptation steps. A flow diagram of Algo.~\ref{alg:mlkp_train} is provided in the Appendix.

\noindent
\textbf{Meta-Training.}~
As presented in Alg.~\ref{alg:mlkp_train}, the meta-training process consists of two main loops: the outer loop over meta-optimization steps (line 2) and the inner loop over tasks (line 6) that constitutes the task adaptation stage.

At each outer loop iteration, a batch of new tasks is sampled from the task distribution (line 4). Each task corresponds to our modified KernelGAN process which consists of an LR image and a kernel \textit{k}, $\mathcal{T}^i = \left< I^{\text{LR},i}, k \right>$. More details on the sampling process for $\mathcal{T}^i$ can be found in the Appendix. 
For each task in the batch, we update the task-specific parameters (line 9), $\theta_G^i$ and $\theta_D^i$, for $N_{\text{adapt}}$ steps, using the following task adaptation loss, which comprises two terms: the L1-norm variant of LSGAN loss~\cite{Mao2017LeastSG} (Eq.~(\ref{eq:lsgan})) and a sum-to-one term $\mathcal{L}^{\text{STO}}(\hat{k}^i)=|1 - \sum_{x,y}\hat{k}^i_{x,y}|$ where $\hat{k}^i$ is the estimated kernel for the $i$-th task.
\begin{small}
    \begin{align}
        \mathcal{L}_G^{\text{LSGAN}}(\theta_D, \theta_G) &= \E_{p^{\text{LR}} \sim patches(I^{\text{LR}})} |D(G(p^{\text{LR}};\theta_G);\theta_D) - 1|  \nonumber \\
        \mathcal{L}_D^{\text{LSGAN}}(\theta_D, \theta_G) &= \E_{p^{\text{LR}} \sim patches(I^{\text{LR}})} \Bigg[ \frac{1}{2}|D(p^{\text{LR}};\theta_D) - 1| \nonumber \\ 
        &+ \frac{1}{2}|D(G(p^{\text{LR}};\theta_G);\theta_D)| \Bigg]
        \label{eq:lsgan}
    \end{align}
\end{small}

\noindent
The $\mathcal{L}^{\text{STO}}$ term encourages the estimated kernel to sum to one, resulting in a Gaussian-like kernel at every adaptation step\footnote{As the kernel is not the output but rather derived from $\theta_G$ (Eq.~(\ref{eq:kernel_deriv})), the use of a softmax layer would not suffice. }.
Finally, we compute the adapted parameters using the following objective functions on task $\mathcal{T}^i$:
\begin{small}
    \begin{align}\label{eq:eq_inner_loop}
        \mathcal{L}_G^{\text{task}}(\theta_D^i, \theta_G^i) &= \mathcal{L}_G^{\text{LSGAN}}(\theta_D^i, \theta_G^i) + \zeta                       \mathcal{L}^{\text{STO}}(\text{DK}(\theta_G^i)) \nonumber \\
        \mathcal{L}_D^{\text{task}}(\theta_D^i, \theta_G^i) &= \mathcal{L}_D^{\text{LSGAN}}(\theta_D^i, \theta_G^i) 
    \end{align} %
\end{small}

\noindent
where $\zeta$ is a hyperparameter and $\text{DK}(\cdot)$ is the kernel decoder from Eq.~(\ref{eq:kernel_deriv}).


In regular intervals during the task adaptation stage, we compute the meta-objective by introducing an additional loss term: $\mathcal{L}^{\text{K{\text -}PIX}}(k, \hat{k})$=$\sum_{x,y}|\hat{k}_{x,y} - k_{x,y}|$ that computes the pixel-wise loss given the estimated task kernel $\hat{k}$. Hence, the covariance constraint is implicitly captured in this supervised learning process.
Overall, our meta-objectives are defined as:
\begin{small}
    \begin{align}\label{eq:eq_outer_loop}
        &\begin{aligned}
             \mathcal{L}_G^{\text{meta}}(\theta_D^i, \theta_G^i, k) = \big[ &\omega \mathcal{L}^{\text{K{\text -}PIX}}(k,\text{DK}(\theta_G^i)) \nonumber \\
                &+ \eta \mathcal{L}_G^{\text{LSGAN}}(\theta_D^i, \theta_G^i) \nonumber \\
                &+ \zeta \mathcal{L}^{\text{STO}}(\text{DK}(\theta_G^i)) \big] \nonumber \\
        \end{aligned}\\
        &\begin{aligned}
        \mathcal{L}_D^{\text{meta}}(\theta_D^i, \theta_G^i) = \mathcal{L}_D^{\text{LSGAN}}(\theta_D^i, \theta_G^i)
        \end{aligned}
    \end{align}
\end{small}
\noindent 
where $\omega$, $\eta$ and $\zeta$ are hyperparameters. Following best practices from MAML++~\cite{Antoniou2019HowTT}, we compute the meta-objective losses for both $G$ and $D$ every $N_{\text{val}}$ inner-loop steps and store in dictionary $b$ (lines 10-12). Backpropagating at not only the last but also the preceding steps has been shown to alleviate gradient instability during the meta-training phase.

After the completion of each task adaptation process, we perform an update of the base parameters using a weighted sum of the recorded meta-objectives\footnote{For the weight values, we follow the schedule presented by~\cite{Soh2020}. We defer the details to the Appendix.} (lines 16-17). This process is repeated for $N_{\text{steps}}$, until the final base parameters, $\hat{\theta}_G$ and $\hat{\theta}_D$, are available to be used for inference.


\begin{table}[t]
\centering
\resizebox{0.48\textwidth}{!}{%
\begin{tabular}{l|l|l|l|l|l} \toprule
\multicolumn{1}{c}{\textbf{Method}} & \multicolumn{1}{c}{\textbf{Steps}} & \multicolumn{1}{c}{\textbf{Set14}} & \multicolumn{1}{c}{\textbf{B100}} & \multicolumn{1}{c}{\textbf{Urban100}} & \multicolumn{1}{c}{\textbf{DIV2K}} \\ \midrule
\multicolumn{6}{c}{$\times$2} \\ \hline
Bicubic & - & 24.82/0.6910/-/- &25.17/0.6603/-/- &22.31/0.6471/-/- &26.94/0.7642/-/-\\
Double-DIP~\cite{Gandelsman2019DoubleDIPUI} + USRNet & 1000
&	21.84/0.5984/41.34/5.96	&	18.52/0.4475/39.99/8.06
&	19.86/0.5827/37.95/5.93		&	24.33/0.7069/37.62/5.06\\
DIP-FKP~\cite{FKP} + USRNet & 1000  &28.90/0.8324/45.32/2.95	&28.64/\textbf{0.8224}/\textbf{46.67}/2.95 &26.00/0.8064/42.59/3.53  &30.31/0.8722/43.19/3.30\\
BSRDM~\cite{BSRDM} + USRNet & 440  & \textbf{29.38}/0.8240/43.77/2.84	& 28.65/0.7936/40.76/3.21 & 26.20/0.8080/44.24/2.70  & 30.68/0.8695/46.00/2.57 \\
MetaKernelGAN (ours) + USRNet & 200 & 28.71/\textbf{0.8331}/\textbf{46.23}/\textbf{2.44} & \textbf{28.95}/0.8222/45.94/\textbf{2.38} & \textbf{26.78}/\textbf{0.8355}/\textbf{47.37}/\textbf{2.16} & \textbf{31.67}/\textbf{0.8944}/\textbf{48.00}/\textbf{2.08} \\
\hdashline
GT + USRNet (upper bound) & - & 32.56/0.8944/-/- & 31.33/0.8771/-/- & 29.96/0.8954/-/- & 34.59/0.9268/-/- \\ 
\hline
\multicolumn{6}{c}{$\times$2 with Non-Gaussian Kernel} \\ \hline
Bicubic & - & 24.80/0.6907/-/- &25.17/0.6609/-/- &22.31/0.6476/-/- &26.93/0.7640/-/- \\
Double-DIP~\cite{Gandelsman2019DoubleDIPUI} + USRNet & 1000
& 21.48/0.5871/40.69/5.99 	&	18.28/0.4367/39.34/8.16
&	19.66/0.5748/37.44/5.96		&	24.12/0.6995/37.14/5.10 \\
DIP-FKP~\cite{FKP} + USRNet & 1000
&	28.18/0.8153/43.30/2.96 	&	27.88/0.8002/\textbf{44.36}/2.87
&	25.31/0.7888/41.23/3.50  &	29.99/0.8646/41.63/3.30 \\
BSRDM~\cite{BSRDM} + USRNet & 440  & \textbf{29.01}/\textbf{0.8155}/41.24/2.85	& 28.41/0.7879/39.77/3.13 & 25.23/0.7875/42.47/2.70  & 29.90/0.8631/43.16/2.58 \\
MetaKernelGAN (ours) + USRNet & 200 & 28.05/0.8153/\textbf{43.68}/\textbf{2.36} & \textbf{28.20}/\textbf{0.8031}/43.69/\textbf{2.32} & \textbf{25.84}/\textbf{0.8122}/\textbf{44.72}/\textbf{2.12} & \textbf{30.96}/\textbf{0.8830}/\textbf{44.96}/\textbf{2.08} \\

\hdashline
GT + USRNet (upper bound) & - & 32.47/0.8968/-/- & 31.34/0.8803/-/- & 30.08/0.8999/-/- & 34.59/0.9272/-/- \\ \hline

\multicolumn{6}{c}{$\times$2 with Image Noise of Level 10 (3.92\%)} \\ \hline
Bicubic & - & 24.69/0.6639/-/- &24.98/0.6321/-/- &22.20/0.6162/-/- &26.65/0.7273/-/- \\
Double-DIP~\cite{Gandelsman2019DoubleDIPUI} + USRNet & 1000
	&	24.32/0.7023/41.29/5.97	&	21.96/0.6010/39.91/8.25
&	20.53/0.6174/37.66/5.73	&   25.46/0.7547/37.43/5.03\\
DIP-FKP~\cite{FKP} + USRNet & 1000
	&	28.36/0.7772/\textbf{44.70}/2.95	&	27.64/\textbf{0.7401}/\textbf{45.22}/3.11
&   25.47/0.7654/42.49/3.58	&   29.49/0.8259/43.10/3.31\\
BSRDM~\cite{BSRDM} + USRNet & 440  & \textbf{28.37}/0.7716/43.23/3.06	& 27.62/0.7308/41.15/3.18 & 25.85/0.7723/\textbf{45.13}/\textbf{2.59}  & \textbf{30.28}/\textbf{0.8325}/\textbf{46.47}/\textbf{2.25} \\
MetaKernelGAN (ours) + USRNet & 200 & 28.23/\textbf{0.7773}/43.99/\textbf{2.76} & \textbf{27.75}/0.7354/43.07/\textbf{2.98} & \textbf{25.86}/\textbf{0.7780}/44.80/2.60 & 30.25/0.8311/44.91/2.64 \\

\hdashline
GT + USRNet (upper bound) & - & 29.57/0.7990/-/- & 28.56/0.7607/-/- & 27.17/0.8114/-/- & 31.27/0.8497/-/- \\ \hline
\multicolumn{6}{c}{$\times$4} \\ \hline
Bicubic & - & 21.15/0.5280/-/- &22.09/0.5119/-/- &19.30/0.4761/-/- &23.20/0.6329/-/-\\
Double-DIP~\cite{Gandelsman2019DoubleDIPUI} + USRNet & 1000
&   20.48/0.5099/53.13/16.65  &	18.77/0.4028/52.26/22.92
&	17.86/0.4375/50.20/18.05	&	21.48/0.5704/46.33/15.30\\
DIP-FKP~\cite{FKP} + USRNet & 1000
&	24.98/0.6707/54.17/16.28	&	\textbf{24.57}/0.6167/53.46/19.10
&	21.65/0.5947/48.57/23.39	&	25.18/0.7177/46.03/26.76\\
BSRDM~\cite{BSRDM} + USRNet & 440  & 24.64/0.6415/48.09/17.10	& 24.55/0.5898/46.44/21.45 & 21.94/0.5868/49.47/18.66  & 26.72/0.7243/51.45/16.35 \\
MetaKernelGAN (ours) + USRNet & 200 & \textbf{25.46}/\textbf{0.6960}/\textbf{63.13}/\textbf{6.54} & 24.36/\textbf{0.6200}/\textbf{59.84}/\textbf{10.84} & \textbf{22.21}/\textbf{0.6375}/\textbf{61.74}/\textbf{9.01} & \textbf{26.99}/\textbf{0.7600}/\textbf{61.78}/\textbf{8.34} \\

\hdashline
GT + USRNet (upper bound) & - & 27.89/0.7498/-/- & 26.92/0.6986/-/- & 24.95/0.7357/-/- & 29.46/0.8069/-/- \\ \hline
\end{tabular}%
}
\caption{Average Image PSNR/Image SSIM/Kernel PSNR/$\mathcal{L}^{\text{K{\text -}COV}}$ of MetaKernelGAN and existing joint learning methods on standard benchmarks across five runs.}

\label{tab:dip_table}
\vspace{1mm}
\end{table}

\noindent
\textbf{Inference.} To adapt to a new image, we initialize $G$ and $D$ with $\hat{\theta}_G$ and $\hat{\theta}_D$ and adapt these parameters using our task adaptation loss as per Eq.~(\ref{eq:eq_inner_loop}). The estimated kernel is derived from MetaKernelGAN's generator as per Eq.~(\ref{eq:kernel_deriv}). Full algorithm is shown in Appendix Alg.~\ref{alg:mlkp_test}.

\begin{table}[t]
\centering
\resizebox{0.4\textwidth}{!}{%
\begin{tabular}{lcccc}
\toprule
{\textbf{Method}} & {\begin{tabular}[c]{@{}c@{}}\textbf{$\times$2 Peak Memory}\\ \textbf{Usage (GB)}\end{tabular}} & {\begin{tabular}[c]{@{}c@{}}\textbf{$\times$4 Peak Memory}\\ \textbf{Usage (GB)}\end{tabular}} & {\begin{tabular}[c]{@{}c@{}}\textbf{$\times$2 Latency}\\ \textbf{(Seconds)}\end{tabular}} & {\begin{tabular}[c]{@{}c@{}}\textbf{$\times$4 Latency}\\ \textbf{(Seconds)}\end{tabular}} \\ 
\midrule
KernelGAN & \phantom{0}0.15 & \phantom{0}0.15 & 133.5 & 109.5 \\ 
KernelGAN-FKP & \phantom{0}0.23 & \phantom{0}0.23 & 113.9 & 441.6 \\ 
Double-DIP & 28.42 & 28.42 & 725.1 & 731.0 \\ 
DIP-FKP & 28.42 & 28.42 & 719.9 & 725.8 \\ 
BSRDM & \phantom{0}9.66 & 19.41 & 101.1 & 200.2 \\ 
DSKernelGAN/MetaKernelGAN & \phantom{0}0.15 & \phantom{0}0.15 & \phantom{00}7.1 & \phantom{00}7.4 \\ 
\bottomrule
\end{tabular}
}
\caption{Inference cost to estimate the kernel from a LR image corresponding to a 1356$\times$2040 HR image.}
\label{tab:cost}
\end{table}

\section{Experiments}\label{sec:experiments}

\noindent
\textbf{Models.}~For MetaKernelGAN, we employ a 6-layer deep linear generator and adopt KernelGAN's discriminator. More details are provided in the Appendix. 


\noindent
\textbf{Training Setup.}~We use the DIV2K~\cite{DIV2K} training set which consists
of images that are captured with diverse cameras. 
We crop the sampled HR image into patches of $192\times192$ for each task. 
The cropped image is then randomly rotated by $90^{\circ}$ and/or flipped vertically and/or horizontally before being downsampled for each task. 
We meta-learn both $\theta_G$ and $\theta_D$, leaving meta-learning $\theta_G$ only as an ablation study in the Appendix.

\noindent
\textbf{Optimization Parameters.}~Our MetaKernelGAN is trained for 100,000 meta-train steps, $N_{\text{steps}}$$=$$10^5$. We use SGD~\cite{Kiefer1952StochasticEO} and Adam~\cite{Kingma_2014} for the task- and meta-optimizer respectively and adopt the first-order MAML~\cite{MAML} algorithm due to the high memory cost of our GAN-based training. 
We set $N_{\text{adapt}}$$=$$25$, and leave experimenting with different $N_{\text{adapt}}$ values in the Appendix. 
We set $N_{\text{val}}$$=$$5$, evaluating the meta-objectives at every 5 steps, \textit{i.e.}~at steps $5,10,15,20,25$. 
We set the discriminator's input patch size to 32 and use the following hyperparameters: {\small$\alpha_G$$=$$0.01$, $\alpha_D$$=$$0.2$, $\beta_G$$=$$1e^{-4}, \beta_D$$=$$1e^{-4}$, $\omega$$=$$1.0$, $\eta$$=$$1.0$}, and {\small$\zeta$$=$$0.5$}. 

\noindent
\textbf{Kernel Distribution.}~Following KernelGAN-FKP, we consider an anistropic Gaussian distribution; each kernel is determined based on a covariance matrix, $\Sigma$, which is parameterized by
a random angle, $\Theta \sim U[0, \pi]$, and two random eigenvalues, $\lambda_1,\lambda_2 \sim U[0.35, 5.0]$.
Similarly to both KernelGAN and KernelGAN-FKP, we use the computed $\times2$ kernel to analytically derive the $\times4$ kernel of size $21\times21$. 

\noindent
\textbf{Evaluation Setup.}~We use the benchmark datasets provided by ~\cite{FKP} for KernelGAN-FKP in all our experiments. Specifically, we use Set14~\cite{Set14}, B100~\cite{B100}, Urban100~\cite{Urban100}, and DIV2K validation set, in which each $I^{\text{HR}}$ image is downsampled using a different randomly generated kernel, $k$, to form $I^{\text{LR}}$. The benchmark also consists of unseen degradations, namely non-Gaussian kernels and noisy images derived by adding noise to $k$ or $I^{\text{LR}}$, respectively. Concretely, uniform multiplicative noise of up to $40\%$ of the maximum kernel pixel value is added to each randomly generated Gaussian kernel, followed by normalization so that each kernel sums to one.
For noisy images, up to $3.92\%$ of the maximum pixel value is added to $I^{\text{LR}}$ at the end of the degradation process. 
Lastly, we utilize USRNet-tiny~\cite{USRNet} as our downstream non-blind SR model and report the upper bound of attainable image performance by using the ground-truth (GT) kernel. 

\noindent
\textbf{Evaluation Metrics.}~We follow previous works and quantify the accuracy of the estimated kernel with respect to its ground truth using PSNR, which captures pixel magnitude.
Unlike previous works, we also examine its covariance by using the sum of absolute distances between the discretized covariance matrices, $\hat{\Sigma}$, of the kernel estimate and its ground truth: {\small\mbox{$\mathcal{L}^{\text{K{\text -}COV}} = \sum^N_{x,y} \big| \hat{\Sigma}^{\text{GT}}_{x,y} - \hat{\Sigma}^{\text{Est}}_{x,y} \big|$}},
where $\hat{\Sigma}^\text{{GT}}$ and $\hat{\Sigma}^\text{{Est}}$ are the covariance matrix of the GT kernel and the estimated kernel respectively. As $\hat{\Sigma}$ is a discrete representation of $\Sigma$, the amount of discretization error depends on the kernel size. 
Details of how we derive $\hat{\Sigma}$ can be found in the Appendix. 
Discussions on the choice of this covariance metric can be found in Section~\ref{sec:limitations}.
For images, we compute PSNR and SSIM on the Y-channel after shaving them by the scale factor.
We report the average score across five runs for all quantitative results.

\begin{figure}[ht!]
    \centering
    \vspace{-3.5mm}
    \includegraphics[ clip,width=0.40\textwidth]{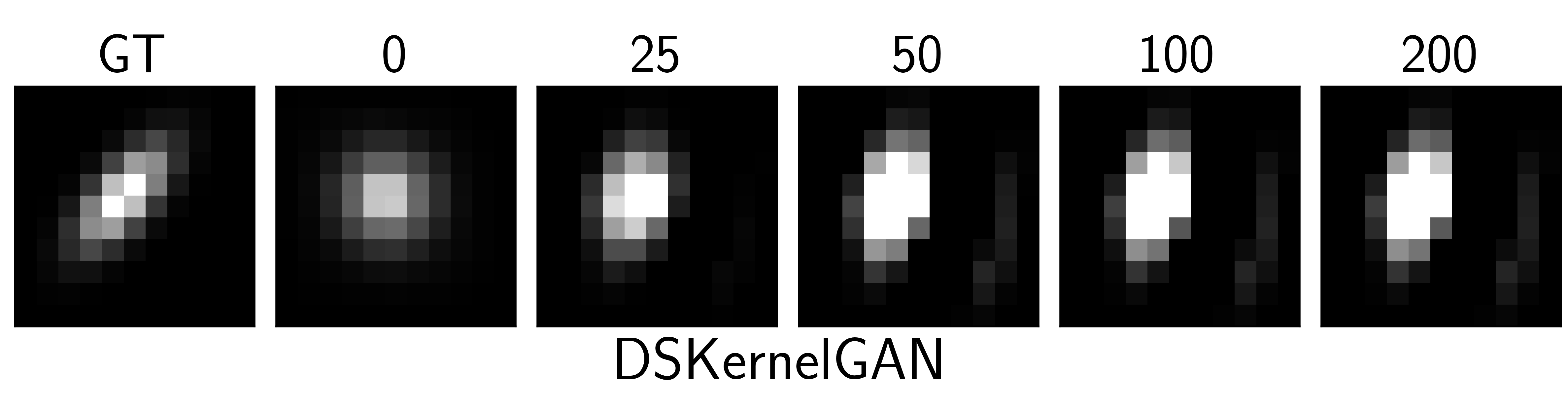}
    \vspace{-1.0mm}
    \includegraphics[ clip,width=0.40\textwidth]{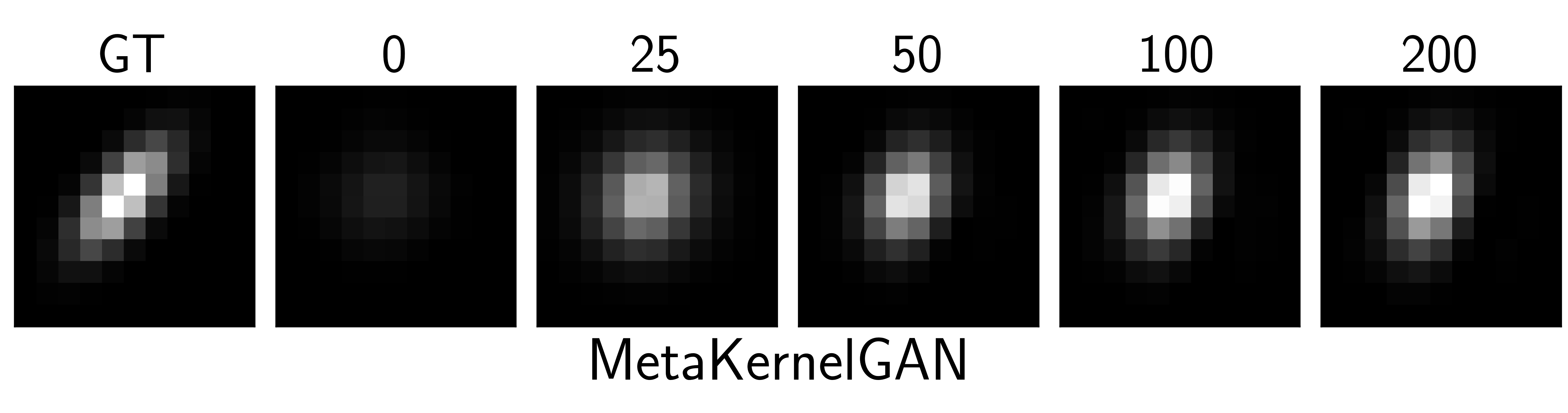}
    \vspace{-1em}
    \caption{Kernel after 25, 50, 100, 200 adaptation steps for $\times$2 upsampling on \textit{108005.png} of B100. More examples in Appendix.
    }
    \label{fig:adaptation_step}
    \vspace{-1.8em}
\end{figure}

\subsection{Comparison with Patch Recurrence Methods}
\label{sec:kg_eval_comparison}

In Table~\ref{tab:kg_table}, we compare our work with other KernelGAN-based approaches, which solely utilize the patch recurrence of the given image during inference. 
We ran each method per-image using the recommended number of steps in each prior work and report both the average image and kernel performance.

\noindent
\textbf{Performance.}~For most natural images, the larger the image, the stronger the patch recurrence~\cite{zontak2011internal}, which can be effectively utilized by KernelGAN and its variants.
Hence, as KernelGAN-FKP requires a larger patch size than KernelGAN, KernelGAN-FKP's performance deteriorates more for smaller spatial resolution benchmarks.
As a baseline, we pretrained KernelGAN in a supervised manner, without meta-learning, optimizing both the generator and discriminator using the meta-objective (Eq~\ref{eq:eq_outer_loop}). We named this approach DataSetKernelGAN (DSKernelGAN) and trained it with the same hyperparameters as MetaKernelGAN's. Surprisingly, DSKernelGAN outperforms previous KernelGAN-based approaches by a large margin, even on unseen distributions. This can be attributed to starting with a reasonable kernel, \textit{i.e.}~the expected value of the training set, as opposed to randomly initializing the kernel in the parameter space (KernelGAN) or the flow-based latent space (KernelGAN-FKP). Nevertheless, unlike MetaKernelGAN which directly learns from the adaptation process, DSKernelGAN is more susceptible to adapt to faulty kernels \textit{e.g.} having non-zero values that are discontinuous as shown in Fig.~\ref{fig:adaptation_step}. 
For $\times4$ upsampling, the LR images are tiny and severely limited in internal information. Hence, KernelGAN and KernelGAN-FKP fail when the given image is below the required minimum spatial resolution due to the need for further downsampling.
Conversely, DSKernelGAN \& MetaKernelGAN supports any spatial resolution by padding the images to the minimum size. 
In this limited setting, the performance gap between DSKernelGAN \& MetaKernelGAN is closer together as both methods start from an informed model initialization and adapt to an isotropic Gaussian-like kernel that performs well on average across multiple tasks. An analysis of the fallback on the initial kernel in tiny images can be found in the Appendix. Benchmark and real-world qualitative results can be found in Fig.~\ref{fig:img_w_ker} and \ref{fig:real_world_qualitative} with more examples in the Appendix.

\noindent
\textbf{Inference Cost.}~Table~\ref{tab:cost} shows the peak memory usage and latency cost for explicit kernel estimation methods during inference. The cost of KernelGAN-based approaches is invariant to the image resolution as it utilizes a fixed-size patch for learning. Due to MetaKernelGAN's kernel adaptation stability, the recommended number of steps is fixed at 200, resulting in a speedup of up to 18.8$\times$ for $\times$2 upsampling and $59.6\times$ for $\times$4 upsampling over previous KernelGAN-based methods.


\begin{figure}
    \centering
    \includegraphics[clip,width=0.47\textwidth]{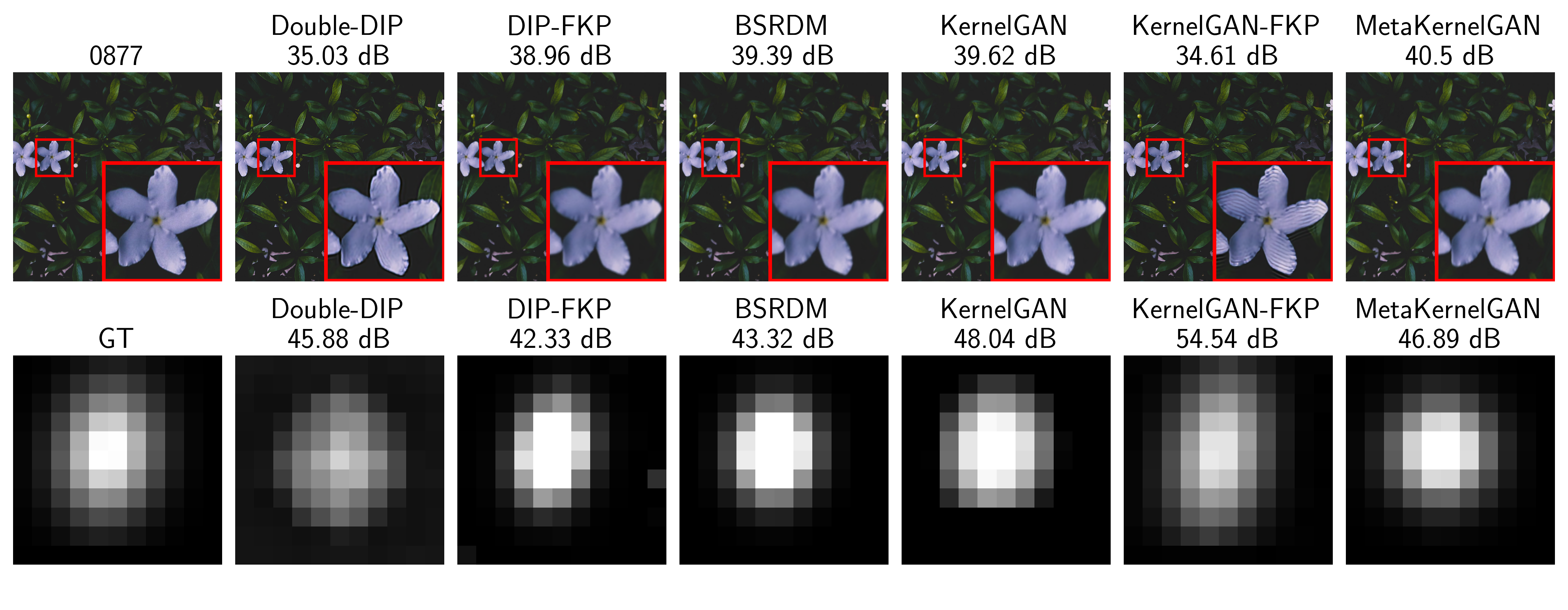}
    \vspace{-1em}
        \caption{Comparison of estimated kernel, along with its image $\times2$ upsampled using USRNet, on \textit{0877.png} DIV2K. Zoom in for best results. See Appendix for more enlarged examples.}
    \label{fig:img_w_ker}
\end{figure}

\subsection{Comparison with Joint Learning Methods}
\label{sec:dip_eval_comparison}

We further compare with DIP-based approaches, which jointly optimize both the image and kernel, and BSRDM~\cite{BSRDM}, which optimizes for the image noise in addition to the image and kernel (Table~\ref{tab:dip_table}). In comparison with patch recurrence methods, these joint learning methods use full-blown models and are significantly more costly in both compute and memory. For a fair comparison with our work, we extract the kernel after optimization and feed it into USRNet for upsampling. 
Due to GPU memory constraints, for DIP-based approaches and BSRDM, we crop $600$$\times$$600$ center image patches to extract $\hat{k}$ for DIV2K. 

\noindent
\textbf{Performance.}~As joint optimization is a more challenging task than patch recurrence approaches, the naive approach, Double-DIP, although effective for image deblurring~\cite{Ren2020NeuralBD}, often fails to find adequate solutions. By restricting the possible set of kernel solutions, DIP-FKP outperforms previous KernelGAN-based approaches by a huge margin. BSRDM took a step further by learning the precision matrix of a Gaussian kernel, hence ensuring that the learned kernel is strictly Gaussian. Additionally, they also explicitly modeled the noise, achieving state-of-the-art for kernel estimation on most benchmarks among joint learning methods. These joint learning methods are known to perform significantly better than patch recurrence methods at a cost of heavy compute and memory.
Despite being orders-of-magnitude more efficient, our approach helps to bridge this performance gap between joint learning methods and patch recurrence methods, having similar performance or even outperforming them on several benchmarks. Notably, our method is able to achieve similar performance with BSRDM without explicitly modeling the noise. Nonetheless, our approach is still limited in cases where images have weak patch recurrence, namely Set14 and B100. This result highlights the potential of better utilizing external images to improve existing unsupervised blind SR methods.



\begin{table}[t]
\centering
\resizebox{0.47\textwidth}{!}{%
\begin{tabular}{l|l|l|l|l} \toprule
\multicolumn{1}{c}{\textbf{Method}} & \multicolumn{1}{c}{\textbf{Set14}} & \multicolumn{1}{c}{\textbf{B100}} & \multicolumn{1}{c}{\textbf{Urban100}} & \multicolumn{1}{c}{\textbf{DIV2K}} \\ \midrule
\multicolumn{5}{c}{$\times$2} \\ \hline
IKC~\cite{IKC} & 27.34/0.8050 & 27.14/0.7712 & 24.64/0.7761 & 29.33/0.8562\\
DAN~\cite{DAN} &  27.06/0.7787 &26.98/0.7481 & 24.39/0.7554 & 29.54/0.8490 \\
DASR~\cite{Wang2021UnsupervisedDR} &  27.34/0.7977 & 27.31/0.7694 & 24.73/0.7811 & 29.74/0.8611\\
MetaKernelGAN (ours) + USRNet &  \textbf{28.71}/\textbf{0.8331} & \textbf{28.95}/\textbf{0.8222} & \textbf{26.78}/\textbf{0.8355} & \textbf{31.67}/\textbf{0.8944} \\


\hline
\multicolumn{5}{c}{$\times$2 with Non-Gaussian Kernel} \\ \hline
IKC~\cite{IKC} &  27.02/0.7765 & 26.82/0.7383 & 24.0/0.7325 & 28.99/0.8357\\
DAN~\cite{DAN} &  26.66/0.7585 & 26.66/0.7277 & 23.84/0.7208 & 28.90/0.8210\\
DASR~\cite{Wang2021UnsupervisedDR} &  26.8/0.7660 & 26.77/0.7344 & 23.97/0.7297 & 29.07/0.8286\\
MetaKernelGAN (ours) + USRNet &  \textbf{28.05}/\textbf{0.8153} & \textbf{28.2}/\textbf{0.8031} & \textbf{25.84}/\textbf{0.8122} & \textbf{30.96}/\textbf{0.8830} \\

\hline

\multicolumn{5}{c}{$\times$2 with Image Noise of Level 10 (3.92\%)} \\ \hline
IKC~\cite{IKC} &  26.02/0.6895& 25.96/0.6524&23.23/0.6364 &27.65/0.7231\\
DAN~\cite{DAN} &  26.02/0.6828& 25.95/0.6500&23.28/0.6363 &27.71/0.7222\\
DASR~\cite{Wang2021UnsupervisedDR} &  26.07/0.6843 & 25.98/0.6506 & 23.32/0.6375  & 27.75/0.7223\\
MetaKernelGAN (ours) + USRNet &  \textbf{28.23}/\textbf{0.7773} & \textbf{27.75}/\textbf{0.7354} & \textbf{25.86}/\textbf{0.7780} & \textbf{30.25}/\textbf{0.8311} \\
\hline
\multicolumn{5}{c}{$\times$4} \\ \hline
IKC~\cite{IKC}& 24.51/0.6640 & 24.22/0.6054 & 21.67/0.5952 & 25.83/0.7235\\
DAN~\cite{DAN}& 24.6/0.6534 & 24.39/0.5917 & 21.69/0.5877 & 26.36/0.7247\\
DASR~\cite{Wang2021UnsupervisedDR} &  25.24/0.6719 & \textbf{24.86}/0.6147 & 22.13/0.6118 & 26.8/0.7404\\
MetaKernelGAN (ours) + USRNet &  \textbf{25.46}/\textbf{0.6960} & 24.36/\textbf{0.6200} & \textbf{22.21}/\textbf{0.6375} & \textbf{26.99}/\textbf{0.7600} \\

\hline
\end{tabular}%
}
\caption{\footnotesize Comparing Image PSNR/SSIM with implicit degradation blind SR methods on standard benchmarks.}

\label{tab:implicit_table}
\vspace{1mm}
\end{table}

\noindent
\textbf{Inference Cost.}~
We use a $1356$$\times$$2040$ HR image to estimate the kernel from its LR image, cropped to a maximum size of $600$$\times$$600$, given the recommended number of steps. MetaKernelGAN achieves a speedup of up to $102.1\times$ and $98.7\times$ for $\times$2 and $\times$4 upsampling over DIP-based methods, respectively (Table~\ref{tab:cost}).
Additionally, DIP-based approaches have a peak memory usage of 28.4GB, two orders of magnitude higher than KernelGAN-based approaches. BSRDM uses a smaller variant of the model used in DIP-based approaches and requires fewer optimization steps. Nonetheless, it is still $14.24\times$ and $27.05\times$ slower and has a peak memory usage of $64.4\times$ and $129.4\times$ more than MetaKernelGAN for $\times$2 and $\times$4 upsampling, respectively.

\subsection{Comparison with Implicit Estimation}
\label{sec:implicit_ker_eval_comparison}

Although our work focuses on explicit kernel estimation, we compare it with several implicit approaches to underline their benefits and limitations. These blind SR methods avoid explicitly estimating the kernel and, instead, learn a degradation representation optimized directly in the image space in a supervised manner. We compare with three prominent works, namely IKC~\cite{IKC}, DAN~\cite{DAN}, and DASR~\cite{Wang2021UnsupervisedDR}, by evaluating the recommended pre-trained model provided by the authors. 

\begin{table}[t]
\centering
\resizebox{0.47\textwidth}{!}{%
\begin{tabular}{l|l|l|l|l|l}
\toprule
\multicolumn{1}{c}{\begin{tabular}[c]{@{}c@{}}\textbf{Method}\\ \textbf{Comparisons}\end{tabular}} & \multicolumn{1}{c}{\begin{tabular}[c]{@{}c@{}}\textbf{Kernel}\\ \textbf{Metrics}\end{tabular}} & \multicolumn{1}{c}{\textbf{Set14}} & \multicolumn{1}{c}{\textbf{B100}} & \multicolumn{1}{c}{\textbf{Urban100}} & \multicolumn{1}{c}{\textbf{DIV2K}} \\ \midrule
\multicolumn{6}{c}{$\times$2} \\ \hline
MetaKernelGAN & PSNR & 0.7157/0.7450 & 0.5558/0.5798 & 0.5161/0.5854 & 0.5189/0.5805 \\ 
\& KernelGAN-FKP & $\mathcal{L}^{\text{K{\text -}COV}}$ & 0.7543/0.8373 & 0.3798/0.6451 & 0.5035/0.6247 & 0.5512/0.4508 \\ \hline
MetaKernelGAN & PSNR & 0.6047/0.8021 & 0.5497/0.6914 & 0.7554/0.7780 & 0.4750/0.6718 \\ 
\& DIP-FKP & $\mathcal{L}^{\text{K{\text -}COV}}$ & 0.5082/0.7362 & 0.4786/0.5734 & 0.4556/0.4709 & 0.4181/0.5365 \\ 
\hline
MetaKernelGAN & PSNR & 0.5137/0.6967 & 0.7092/0.8014 & 0.7349/0.7919 & 0.5292/0.7935 \\ 
\& BSRDM & $\mathcal{L}^{\text{K{\text -}COV}}$ & 0.5991/0.6132 & 0.5463/0.6486 & 0.4792/0.4375 & 0.6953/0.7483 \\
\hline

\multicolumn{6}{c}{$\times$4} \\ \hline
MetaKernelGAN & {PSNR} & {-} & {-} & {-} & 0.5511/0.6820 \\ 
\& KernelGAN-FKP & {$\mathcal{L}^{\text{K{\text -}COV}}$} & {-} & {-} & {-} & 0.6459/0.6975 \\ \hline
MetaKernelGAN & {PSNR} & {0.5496/0.6219} & {0.4537/0.6082} & {0.6984/0.6914} & 0.6506/0.7227 \\ 
\& DIP-FKP & $\mathcal{L}^{\text{K{\text -}COV}}$ & 0.3643/0.5252 & 0.4428/0.5210 & {0.5950/0.5808} & 0.5264/0.6077 \\ 
\hline
MetaKernelGAN & PSNR & 0.4342/0.3363 & 0.5171/0.6208 & 0.6933/0.7420 & 0.6499/0.7895 \\ 
\& BSRDM & $\mathcal{L}^{\text{K{\text -}COV}}$ & 0.2180/0.1604 & 0.5418/0.5508 & 0.6344/0.5832 & 0.5828/0.7059 \\
\bottomrule
\end{tabular}

}
\caption{\footnotesize \mbox{Pearson-\textit{r}}/Spearman-$\rho$ correlation between the kernel metrics and Image PSNR on the outputs of MetaKernelGAN and previous state-of-the-art kernel estimation methods.}

\label{tab:correlation_table}

\end{table}

The image performance of these implicit degradation estimation approaches is stellar, in some cases reaching a similar performance as with the use of the ground-truth kernel. However, as they do not adapt to the given image during inference, their performance drops when the test distribution differs from the training distribution, \textit{e.g.}~\cite{DAN} retrains DAN for different distributions. 
As a result, the performance of these methods is lagging (Table~\ref{tab:implicit_table}), especially in our unseen test distributions, namely non-Gaussian kernels and noisy images, where they perform similarly to one another.
Specifically, IKC and DASR~$\times$2 were trained on isotropic kernels and DAN was trained on a narrower kernel distribution. The closest matching distribution to our evaluation set is DASR~$\times$4, which was trained on anistropic kernels with similar widths ($\lambda_1,\lambda_2 \sim U[0.2, 4.0]$), resulting in a closer performance gap. Hence, implicit degradation methods are ideal for \textit{known} degradation distributions, as they do not require training during inference. Nevertheless, this assumption often does not hold for real-world images in the wild. In terms of memory, IKC, DAN, and DASR have 5.2M, 4.33M, and 5.8M parameters as opposed to MetaKernelGAN's 0.116M.

\subsection{Kernel Analysis}\label{sec:kernel_analysis}

\noindent
\textbf{Correlation between Kernel and Image Fidelity.}~In previous explicit degradation estimation works, kernels are evaluated quantitatively using the PSNR metric. In our experiments, we observe that a higher kernel PSNR does not necessarily imply a higher image fidelity. 
For instance, KernelGAN-FKP achieves superior kernel PSNR for $\times2$ upsampling on DIV2K, but is lacking in image PSNR. 
To elucidate this phenomenon, we measure the correlation between the gain in \textit{i)}~image PSNR and kernel PSNR, and \textit{ii)}~image PSNR and $\mathcal{L}^{\text{K{\text -}COV}}$ between MetaKernelGAN and previous methods and show them in Table~\ref{tab:correlation_table}. 
The corresponding \mbox{Pearson-\textit{r}} and Spearman-$\rho$ correlation coefficients indicate that both the pixel magnitude and the covariance of the estimated kernel are important to evaluate the kernel mismatch for image SR.
As such, the approach of previous kernel estimation works~\cite{KernelGAN,FKP,Ren2020NeuralBD} that solely evaluates the kernel PSNR to indicate the downstream SR performance does not suffice.


To illustrate the cases where the kernel PSNR is lower but the image PSNR is higher, Fig.~\ref{fig:img_w_ker} shows a super-resolved image together with the estimated kernel. 
As MetaKernelGAN is learnt using the GT kernels, it puts greater emphasis on pixels closer to the center, thus underestimating kernels that are slightly sharper than the GT and leading to images that may be slightly blurry. 
In contrast, although KernelGAN-FKP's kernels look closer to the ground truth, it tends to significantly overestimate kernels, yielding smoother kernels and hence overly sharp images and unpleasant artifacts.
Although both under- and over-estimation of kernels are detrimental to the downstream image quality~\cite{Efrat2013}, we argue that undesirable artifacts are less perceptually appealing to the human eye than a slight blur.
Joint learning methods, on the other hand, estimate kernels whose covariance is closer to that of the ground truth than KernelGAN-FKP's. However, its performance on kernel pixel magnitude is still lagging (Table~\ref{tab:dip_table}).  
Moreover, we observe that for some images, DIP-FKP and KernelGAN-FKP produce kernels that have ill-conditioned Gaussian covariance matrices, aggravating kernel mismatch and image SR.
\vspace{-1em}
\begin{figure}[ht!]
 \centering
    \includegraphics[width=0.47\textwidth]{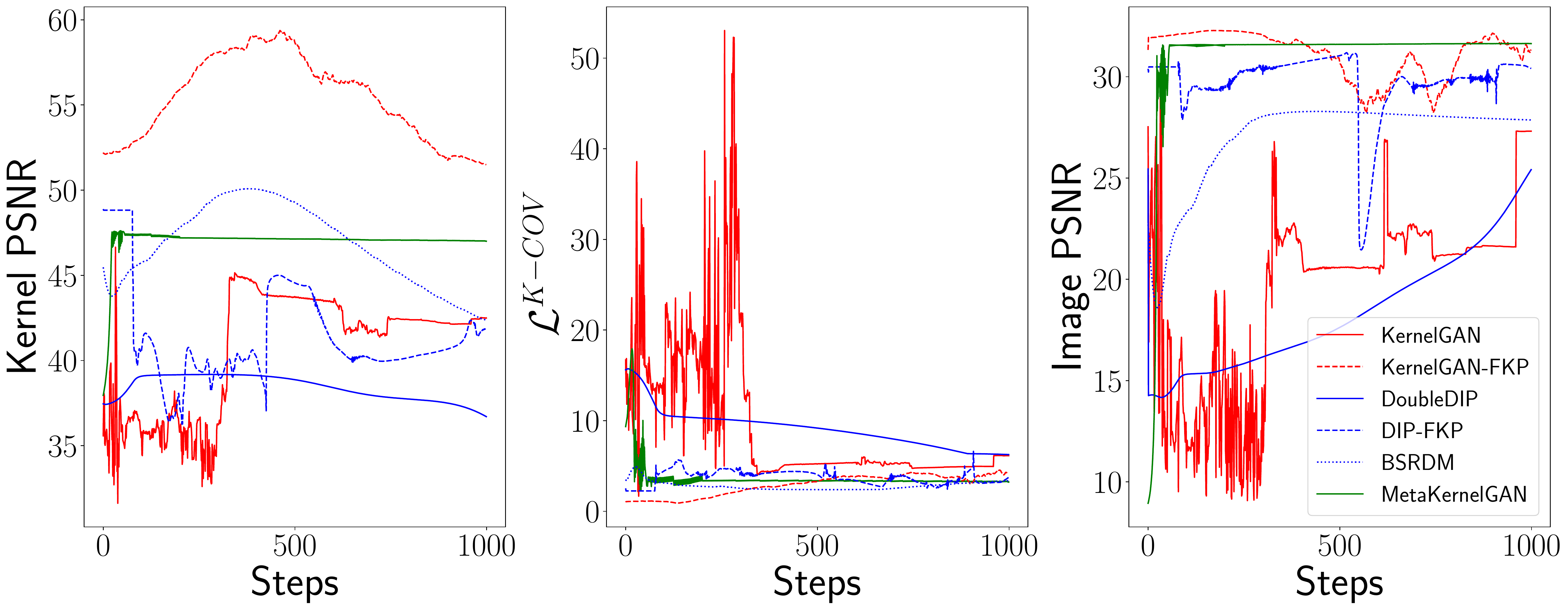}
    \vspace{-1em}
    \caption{Intermediate kernel results and the corresponding downstream image results generated by USRNet for every adaptation iteration ($\times2$ upsampling for \textit{0839.png} in DIV2K).}
    \vspace{-1em}
    \label{fig:convergence_plot}
\end{figure}

\noindent
\textbf{Adaptation Stability.}~Fig.~\ref{fig:convergence_plot} depicts the convergence behavior of MetaKernelGAN compared to the kernel estimation baselines across adaptation steps. 
As DIP-FKP and KernelGAN-FKP optimize directly on the learned kernel manifold, they start off with a Gaussian-like kernel, whereas MetaKernelGAN fast adapts to one within the first $25$ steps.
Furthermore, although KernelGAN-FKP has a more stable convergence than KernelGAN, both methods' performance can vary drastically depending on the number of update iterations and the given image.
Similarly, despite explicitly optimizing for a Gaussian kernel, BSRDM's performance varies across update iterations. ~In contrast, MetaKernelGAN's learned adaptation process results in subtle changes in performance and smoother convergence across iterations. 
This results in a more stable image fidelity result across iterations when the kernels are used by the downstream SR model, underlining the benefits of utilizing external images for the internal learning of a single image.

\begin{figure}[ht!]
    \centering
    \vspace{-2mm}
    \begin{subfigure}[t]{0.47\textwidth}
    \begin{center}
    \includegraphics[ clip,width=1.0\textwidth]{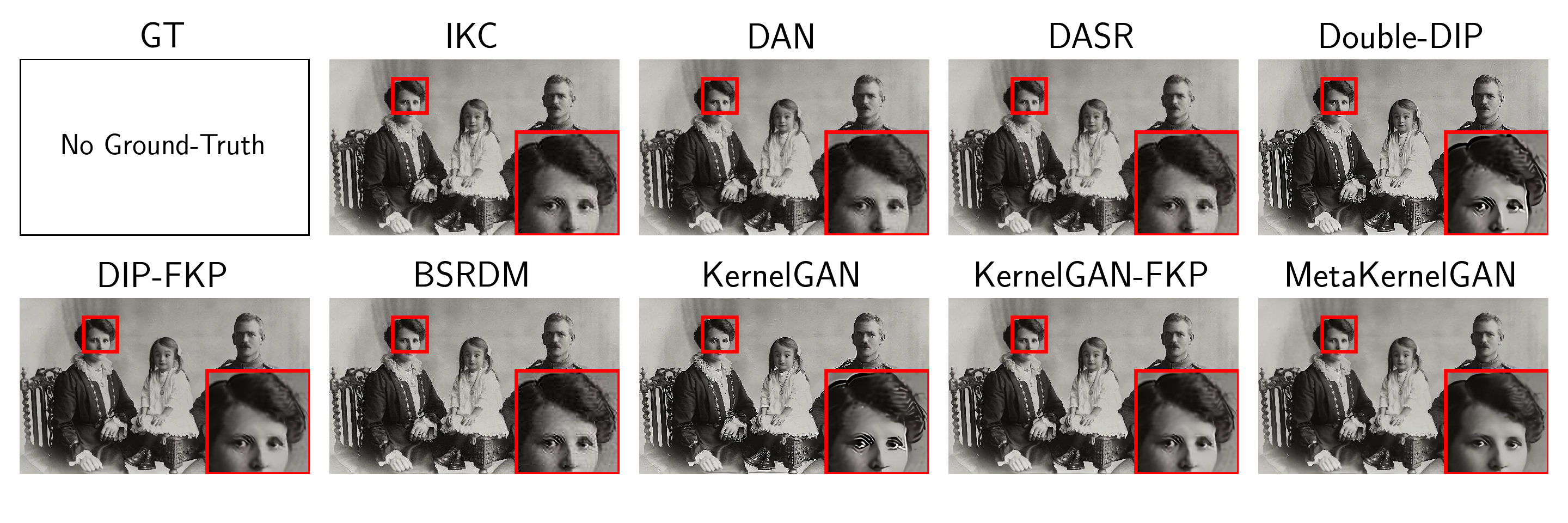}
    \end{center}
    \hfill
    \end{subfigure}
    \vspace{-2em}
    \caption{Real-world visual quality comparison on $\times4$ upsampling. Zoom in for best results. More examples in Appendix.}
    \vspace{-1em}
    \label{fig:real_world_qualitative}
\end{figure}

\section{Discussion}\label{sec:limitations}


\noindent
\textbf{MetaKernelGAN's Use Cases.}
Although MetaKernelGAN's task is kernel estimation, it can be deployed in conjunction with an SR model to perform image SR with improved efficiency and quality. To perform SR, MetaKernelGAN first estimates the kernel of the LR image and then passes it downstream to a non-blind SR model (USRNet in our paper) to produce the HR image.

\noindent \textit{Efficiency Benefits.}~In real-world use-cases, SR models are deployed in two main scenarios: \textit{i)}~on a remote server as-a-service, or \textit{ii)}~on a user device, such as a TV or smartphone. In both cases, reducing the \textit{energy consumption} is of critical importance: on the server side, energy bills contribute the most to the monetary cost, while on the device side, excessive energy consumption affects the battery life. Furthermore, minimizing the carbon footprint is a vital necessity.

The amount of inference-time computation directly determines the energy consumption. As such, with MetaKernelGAN providing significantly faster inference than all existing methods (Table~\ref{tab:cost}), we substantially reduce resource usage, latency and energy, while still delivering accurate kernels. As such, MetaKernelGAN constitutes an enabling component for energy-efficient and high-quality SR.


\noindent
\textbf{Limitations \& Further Directions.}
Despite our substantial gains, we are still faced with a few remaining challenges.

\noindent \textit{Kernel Evaluation}.
We showed that by considering both our proposed $\mathcal{L}^{\text{K-COV}}$ and the status-quo metric (PSNR) of evaluating kernels, we can better reason about the performance of the downstream SR task. 
Nonetheless, these kernel metrics are less effective as the estimated kernel differs more from being Gaussian-like.
As covariance matrices lie on a Riemannian manifold and not in Euclidean space, measuring the geodesic distance instead of the Euclidean distance between two covariance matrices would be a better option~\cite{arsigny2007geometric}.
However, as the learned kernel estimates do not strictly follow the underlying assumed Gaussian distributions, apart from kernels estimated by BSRDM, their covariance matrices might not be positive semi-definite, as often seen in the kernel estimates of DIP-FKP and KernelGAN-FKP.
Similarly, although there are theoretical works that show how the kernel affects the SR result~\cite{Efrat2013}, these cannot be effectively applied to evaluate ill-conditioned covariance matrices, leading to our adoption of $\mathcal{L}^{\text{K-COV}}$ as an approximation. 

On the other hand, it is arguable whether the estimated kernels need to be strictly Gaussian in real-world scenarios. For example, unlike BSRDM, MetaKernelGAN does not explicitly optimize for image noise or restrict the kernel to be Gaussian (Eq.~(\ref{eq:eq_inner_loop})) while achieving similar performance to BSRDM in the image noise experiments. Moreover, as the degradation process of SR is linear (Eq.~(\ref{eq:sr})), the deep linear generator used in MetaKernelGAN can be learnt to represent both the kernel and the noise.


\noindent \textit{Weak Patch Recurrence}.
Another challenge resides in the method we adopted for internal learning; KernelGAN performs poorly on images with weak patch recurrence. Although our approach significantly mitigates this over previous approaches by utilizing an isotropic Gaussian-like kernel, there is still room for improvement in small images. 

\noindent \textit{Optimal Adaptation Steps}.
Although MetaKernelGAN alleviates the kernel adaptation instability found in previous kernel estimation methods, the optimal number of update iterations is still dependent on the given image at test time as it is unsupervised during inference. Nonetheless, we empirically observe that our method requires significantly fewer adaptation steps compared to previous works.


\vspace{-0.8em}
\section{Conclusion}\label{sec:conclusion}
\vspace{-0.5em}
Our work presents the first effective approach for integrating the benefits of learning from a dataset of images in a supervised manner and the unsupervised internal learning of a single image for kernel estimation.
By learning the unsupervised adaptation process across diverse images, MetaKernelGAN robustly estimates accurate kernels applicable to high-fidelity downstream SR tasks, while being an order-of-magnitude faster than previous kernel estimation methods at no extra memory cost.
Unifying both meta-learning and kernel estimation comes at a cost of creating new challenges not present in either of the original algorithms. In contrast to MAML which targets a purely supervised setting, our problem required blending supervised and unsupervised learning and their disparate objectives, as well as meta-learning both generator and discriminator.
We hope that future works would also take advantage of both supervised and unsupervised methodologies such that external images can be used to effectively mitigate existing limitations and better utilize unsupervised methods.


\vspace{-0.3em}
\section*{Acknowledgements}
\vspace{-0.3em}
This work was supported by Samsung AI and the European
Research Council via the REDIAL project.

{\small
\bibliographystyle{ieee_fullname}
\bibliography{ref}
}

\clearpage
\appendix
\interfootnotelinepenalty=10000

\section{Background: Kernel Mismatch}
In real-world settings, the degradation process can often be complex, involving multiple stages of blurring, downsampling and noise addition. 
For cases in which this process is not known, super-resolution models are required to be robust to unknown degradations, \textit{i.e.} they need to be able to upsample any natural image in the wild. 
To this end, Efrat \textit{et al.}\cite{Efrat2013} first shed light on the kernel mismatch phenomenon: how the super-resolved image would be impacted if the estimated kernel differs from the ground-truth kernel, which is assumed to be Gaussian, regardless of the given prior. Specifically, their study demonstrated that a smoother kernel relative to its corresponding ground-truth kernel leads to sharper images while a sharper kernel leads to blurry images. Hence, accurately estimating the ground-truth kernel is crucial in order for the downstream non-blind SR model to produce visually pleasing super-resolved images.

\section{MetaKernelGAN Details}
\noindent
\textbf{Models.}~For our generator, we employ a deep linear architecture with six convolutional layers and without non-linear activations. The layers have kernel sizes of $[7,3,3,1,1,1]$, a stride of $2$ for the last layer and $1$ for the earlier layers, representing the linear operation of applying a blur kernel of size $11\times11$ followed by the subsampling operation. For our discriminator, we adopt KernelGAN's architecture: a 7-layer discriminator with kernel sizes of $[7,1,1,1,1,1,1]$, each followed by a Spectral normalization~\cite{Miyato2018SpectralNF}, batch normalization~\cite{BN}, and ReLU activation, except for the last layer which consists of a sigmoid activation at the end.

\noindent
\textbf{Weighting the Interval Loss Optimization Steps.}~Following MZSR~\cite{Soh2020}, we weight each interval loss optimization step equally at the beginning, then slowly decaying the preceding adaptation steps and converging the weight to the last adaptation step:
\begin{small}
    \begin{flalign}
    &\mathbf{w} \leftarrow \text{GetIntervalLossWeights}(j) \qquad \text{where} \nonumber \\
     &\textbf{w}[0], \cdots, \textbf{w}[N_{\text{val}} - 2]] = \nonumber \\ 
     &\qquad \min(\frac{1}{N_{\text{val}}} - j \cdot \frac{3}{N_{\text{val}} \cdot 10000}, \frac{0.03}{N_{\text{val}}}) \nonumber \\
    &\textbf{w}[N_{\text{val}} - 1] = 1 - \sum_{m=0}^{N_{\text{val}} - 2} w_{m} \nonumber
    \end{flalign} 
\end{small}
where $j \in [1,N_{\text{steps}}]$ is the meta-objective step and $N_{\text{val}}$ is the total number of interval loss evaluation for each task, as used in Alg.~1. 

\noindent
\textbf{Inference Algorithm.}~To adapt the meta-learned GAN on a new LR image (Alg.~\ref{alg:mlkp_test}), we initialize $G$ and $D$ with the meta-learned base parameters, $\hat{\theta}_G$ and $\hat{\theta}_D$ (lines 1-2). We then adapt these parameters using our task adaptation loss as per Eq.~(\ref{eq:eq_inner_loop}) for $N_{\text{adapt}}$ steps (line~3) and the estimated kernel is derived from MetaKernelGAN's generator as per Eq.~(\ref{eq:kernel_deriv}) (line 5). 

\setlength{\textfloatsep}{0pt}
\SetArgSty{textnormal}
\begin{algorithm}[!t]
    \scriptsize
    \SetAlgoLined
    \LinesNumbered
    \DontPrintSemicolon
    \KwIn{Input image $I^{\text{LR}}$}
    \nonl
    \myinput{Number of steps $N_{\text{adapt}}$}
    \nonl
    \myinput{Meta-learned $\hat{\theta}_G$ and $\hat{\theta}_D$}
    \KwOut{Kernel estimate $k^*$ for the given image}
    
    $\theta_G \leftarrow \hat{\theta}_G$, $\theta_D \leftarrow \hat{\theta}_D$  \Comment{Initialize $G$ and $D$ with meta-learned parameters} \\
    \For(\Comment{Adaptation steps over the given image $I^{\text{LR}}$}){$l$ in $[1,N_{\text{adapt}}]$}{
        Compute adapted parameters 
        $\left(\text{Eq.}~(4)\right)$: \newline {$\theta_G \leftarrow \theta_G - \alpha_G \nabla_{\theta_G} \mathcal{L}_G^{\text{task}}(\theta_D, \theta_G)$} 
        \newline
        {$\theta_D \leftarrow \theta_D - \alpha_D \nabla_{\theta_D} \mathcal{L}_G^{\text{task}}(\theta_D, \theta_G)$}
    }
    $\hat{k} \leftarrow \text{DK}(\theta_G)$ \Comment{Derive kernel estimate} \\
    \caption{\footnotesize Inference of MetaKernelGAN}
    \label{alg:mlkp_test}
\end{algorithm}

\subsection{MetaKernelGAN Flow}

\begin{figure}[ht!]
 \centering
        \includegraphics[width=0.5\textwidth]{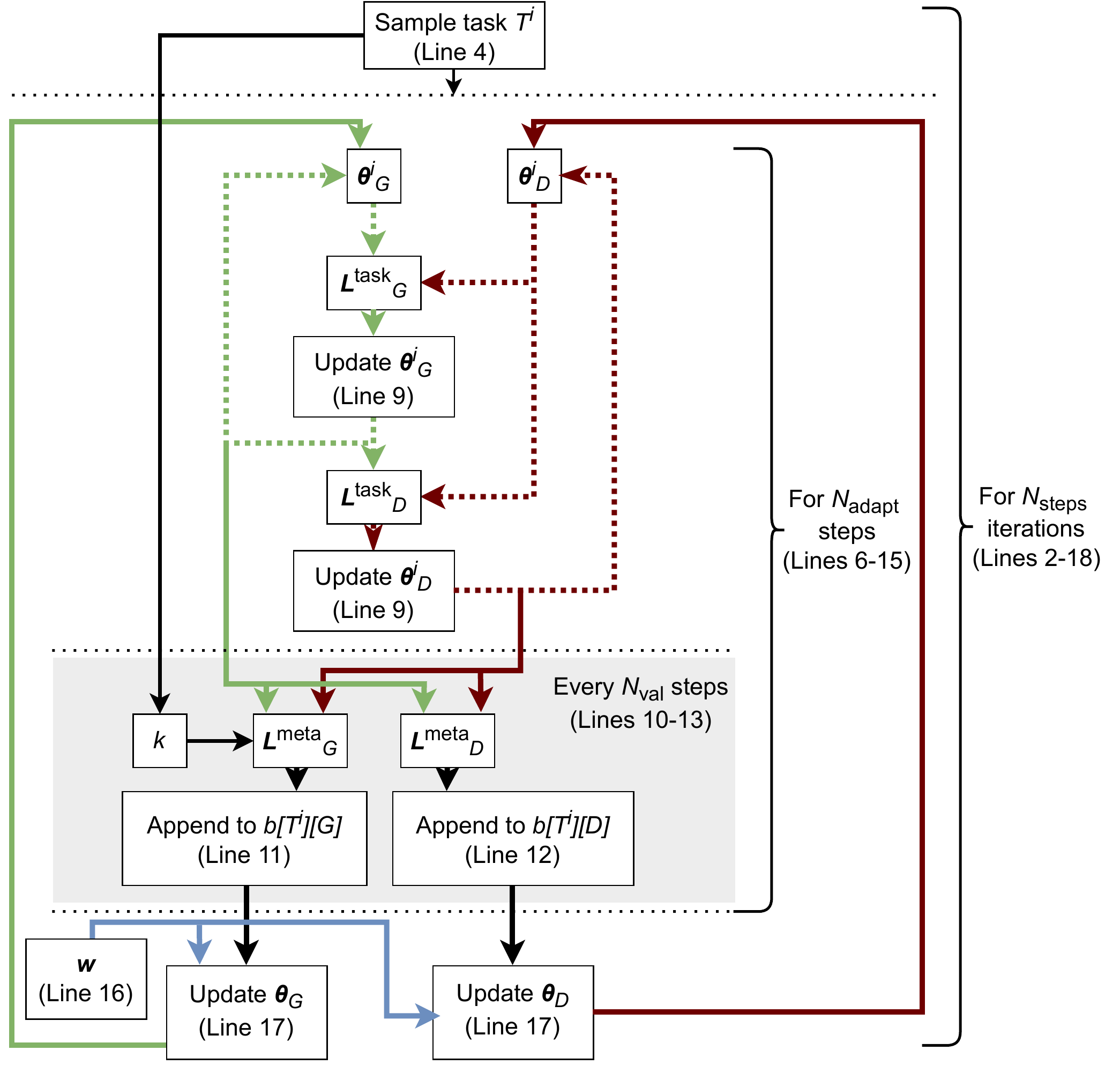}
    \caption{Flow chart of Algo.~\ref{alg:mlkp_train}.}
    \label{fig:flow_chart}
\end{figure}

Fig.~\ref{fig:flow_chart} shows the flow diagram of MetaKernelGAN meta-training stage shown in Algo.~\ref{alg:mlkp_train}, which is described in Section.~\ref{sec:metakernelgan}. Green \& red lines represent updates to the generator the discriminator respectively, with dotted lines representing the task adaptation stage and solid lines representing the meta-optimization stage. The blue lines represent the weighting of the meta-objectives. 


\subsection{MetaKernelGAN Task Design}
The task-related component of our framework comprise: \textit{i)}~the tasks $\mathcal{T}$, and \textit{ii)}~the task probability distribution $p(\mathcal{T})$ that determines the strategy of sampling a task from the candidate tasks, \textit{i.e.}~$\mathcal{T}^i \sim p(\mathcal{T})$.

\noindent
\textbf{Task $\mathcal{T}^i$.}~Each task $\mathcal{T}^i = \left< I^{\text{LR},i}, k \right>$ consists of an LR image and a blur kernel $k$. The LR image is formed in three steps: \textit{i)}~an HR image, $I^{\text{HR},i}$, is first randomly sampled from the given dataset; \textit{ii)}~random augmentation operations, such as flipping or rotation, are applied on the selected HR image; and, finally, \textit{iii)}~an LR image is obtained by applying kernel $k$ on the HR image.
Key to our task setup is that each task encapsulates both a \textit{support} and a \textit{query} set, by containing \textit{multiple} patches with the \textit{same} blur kernel. We denote the support/query set as 
\mbox{$\mathcal{D}^i_s \cup \mathcal{D}^i_q = \{(p^{\text{LR}}, k)\} \quad \exists p^{\text{LR}}$$\in$$patches(I^{\text{LR},i})$},
\textit{i.e.}~the support ($\mathcal{D}^i_s$) and query ($\mathcal{D}^i_q$) sets consist of different patches from the \textit{same} LR image.
With this formulation, the meta-learning method can learn from multiple patches that have been degraded with the same kernel, exploiting in this manner the internal patch recurrence of the given image.
The complete set of tasks $\mathcal{T}$ is defined using a dataset of HR images and a distribution of blur kernels.

\noindent
\textbf{Task Probability Distribution $p(\mathcal{T})$.}~The distribution $p(\mathcal{T})$ controls our task \textit{sampling strategy} and encompasses both the blur kernel and the HR image selection.
First, the kernel $k$ is chosen by \textit{uniformly} sampling from a predefined distribution, \textit{e.g.}~an anisotropic Gaussion distribution. Similarly, the HR image is \textit{uniformly} sampled from the given training dataset. After the sampling process, the final task is constructed by applying $k$ to an augmented version of the HR image to produce its LR counterpart, $I^{\text{LR},i}$.

\noindent
\textbf{Patch Sampling Strategy.}
To evaluate the $\mathcal{L}^\text{LSGAN}$ term used by both the task adaptation loss and the meta-objective of our method, patches are sampled from a task's LR image. To sample an image patch, $p^{\text{LR},i}$, we follow a similar strategy as KernelGAN: we assign a selection probability to each patch in $I^{\text{LR},i}$ based on its gradient magnitude. 
In this manner, patches with higher gradient magnitude have a higher probability to be selected. The rationale behind this strategy is that using flat patches, \textit{i.e.}~with low gradient magnitude, would aggravate the ill-posedness of the kernel estimation problem, leading to a typical isotropic Gaussian kernel~\cite{Liang2021MutualAN}.
Specifically, KernelGAN utilizes the gradient magnitude of $I^{\text{LR},i}$ and its bicubic upsampled super-resolved image to determine each patch's selection probability for the discriminator and generator respectively. Unlike KernelGAN, we utilize the gradient magnitude of $I^{\text{LR},i}$ for the generator instead and take the top-left sub-patch for the discriminator. This empirically results in a slight boost in performance possibly because the discriminator can learn to discriminate between two patches from the same region in the image, as opposed to two randomly sampled regions as implemented in KernelGAN.





\section{Evaluation \& Reproducibility Details}

\noindent
\textbf{Hyperparameter Details.}~We use a task batch size of $1$.\ We divide $\alpha_G$ by $10$ after $50$ \& $200$ adaptation steps during inference. For $\alpha_G$ and $\alpha_D$, we tried values [$0.1$, $0.2$, $0.5$, $0.01$, $0.02$, $0.05$] and picked the ones with the highest performance: $\alpha_G=0.01, \alpha_D=0.2$. 

\noindent
\textbf{Implementation Details.}~MetaKernelGAN was built on top of PyTorch v1.7 and \textit{learn2learn}~\cite{Arnold2020-ss}, an open-source meta-learning framework which we extended to support MetaKernelGAN's components and algorithms. We further integrated part of the code of FKP~\cite{FKP}, MZSR~\cite{Soh2020}, KernelGAN~\cite{KernelGAN}, and USRNet~\cite{USRNet}. 

\noindent
\textbf{Baseline Details.}~Following KernelGAN-FKP, we replace one autoencoder in Double-DIP with a fully-connected network to model the kernel. All prior work results are reported using the associated codebases\footnote{
{\scriptsize\url{https://github.com/JingyunLiang/FKP/}}\\
{\scriptsize\url{https://github.com/greatlog/DAN}}\\
{\scriptsize\url{https://github.com/yuanjunchai/IKC}}\\
{\scriptsize\url{https://github.com/ShuhangGu/DASR}}\\
{\scriptsize\url{https://github.com/zsyOAOA/BSRDM}}\\
}. 

\noindent
\textbf{Subpixel Alignments \& Kernel Shifts.}
We shift the evaluation set of LR images by shifting its blur kernel following ~\cite{USRNet} when evaluating all explicit kernel estimation approaches, including MetaKernelGAN. However, the assumed center of mass of the kernel is slightly different in previous implicit degradation estimation works. Hence, to avoid subpixel misalignments in these cases, we regenerate the evaluation set of LR images by shifting its kernel following ~\cite{KernelGAN} when evaluating IKC, DAN, and DASR.

\noindent
\textbf{Difference in Performance for DIP-FKP.}
The original FKP work uses a \textit{different} kernel distribution to evaluate DIP-FKP and KernelGAN-FKP. For fair comparison, we use the same distribution when evaluating all previous explicit kernel estimation works. Specifically, we adopt the kernel distribution originally proposed for KernelGAN-FKP.

\noindent
\textbf{Covariance of Estimated Kernel.}
We derive the discretized kernel $\hat{k}$ from Eq.~(2) as an {\footnotesize$m$ $\times$ $m$} matrix and calculate the covariance matrix as 
{\small$\hat{\Sigma}$ $=$} $\text{\scriptsize$\begin{bmatrix} a & c \\ c & b \end{bmatrix}$}$ where {\footnotesize$a$ $=$ 
$\text{Var}(col_{\hat{k}})$}, \mbox{\footnotesize$b$ $=$ $\text{Var}(rows_{\hat{k}})$} and 
{\footnotesize$c$ $=$ $\text{Covar}(col_{\hat{k}}, rows_{\hat{k}})$}.


\section{Ablation}

\begin{figure}[ht!]
 \centering
    \includegraphics[width=0.47\textwidth]{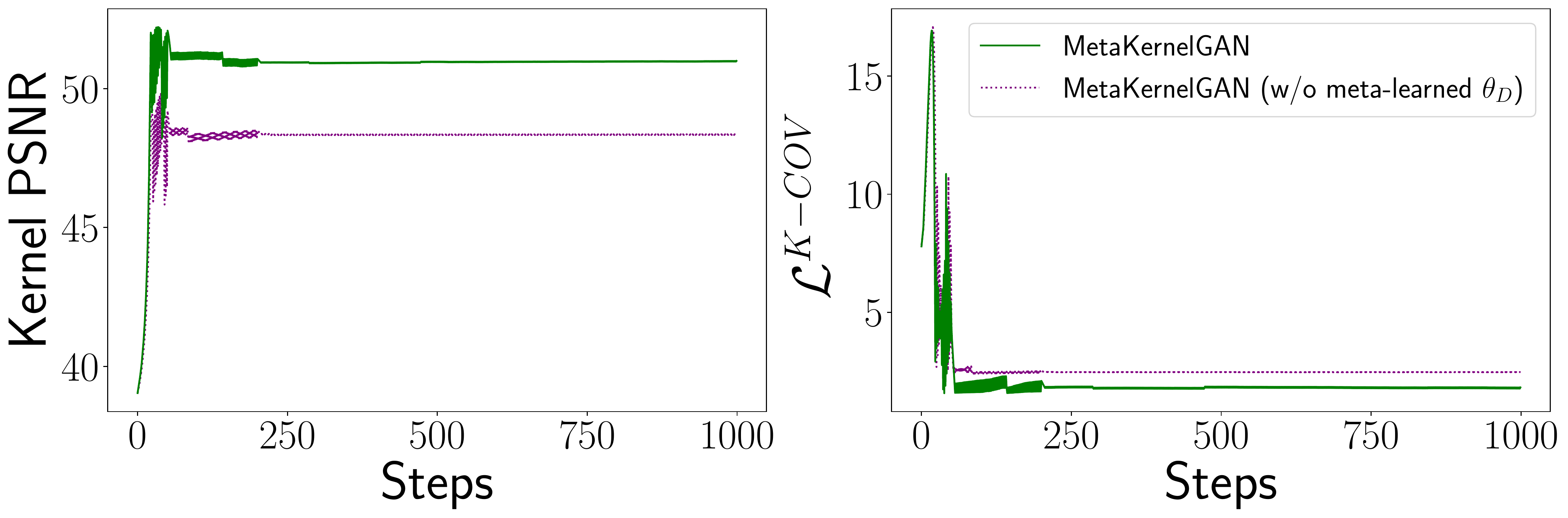}
    \hfill
    \includegraphics[width=0.47\textwidth]{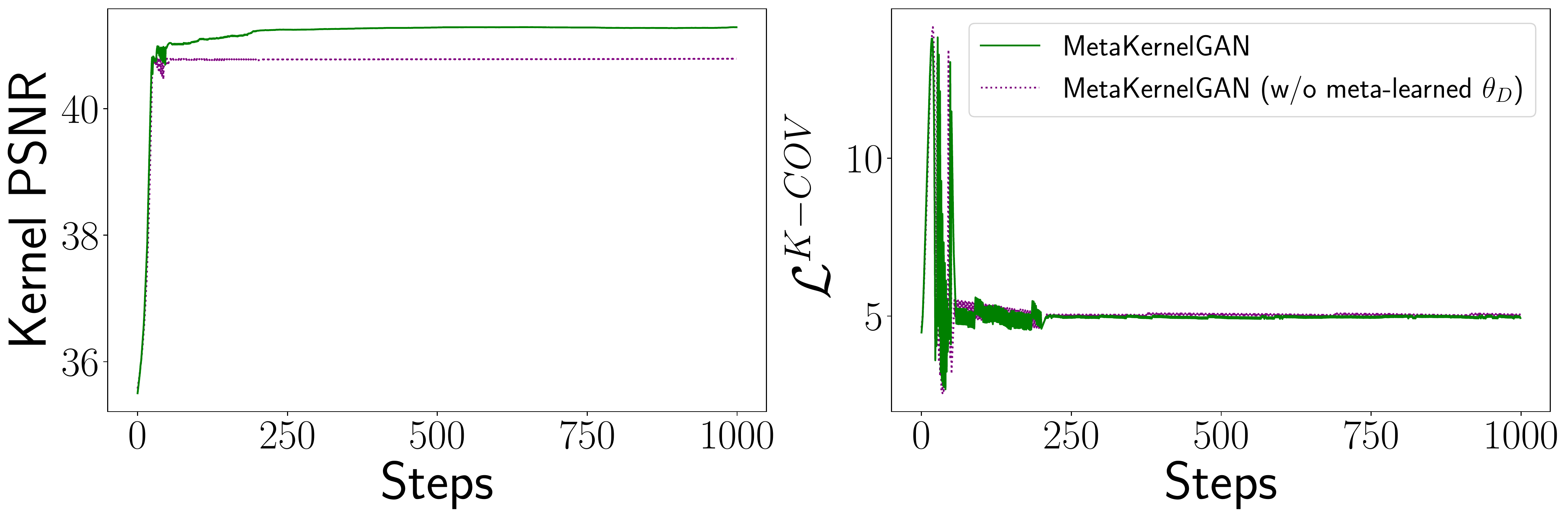}
    \hfill
    \includegraphics[width=0.47\textwidth]{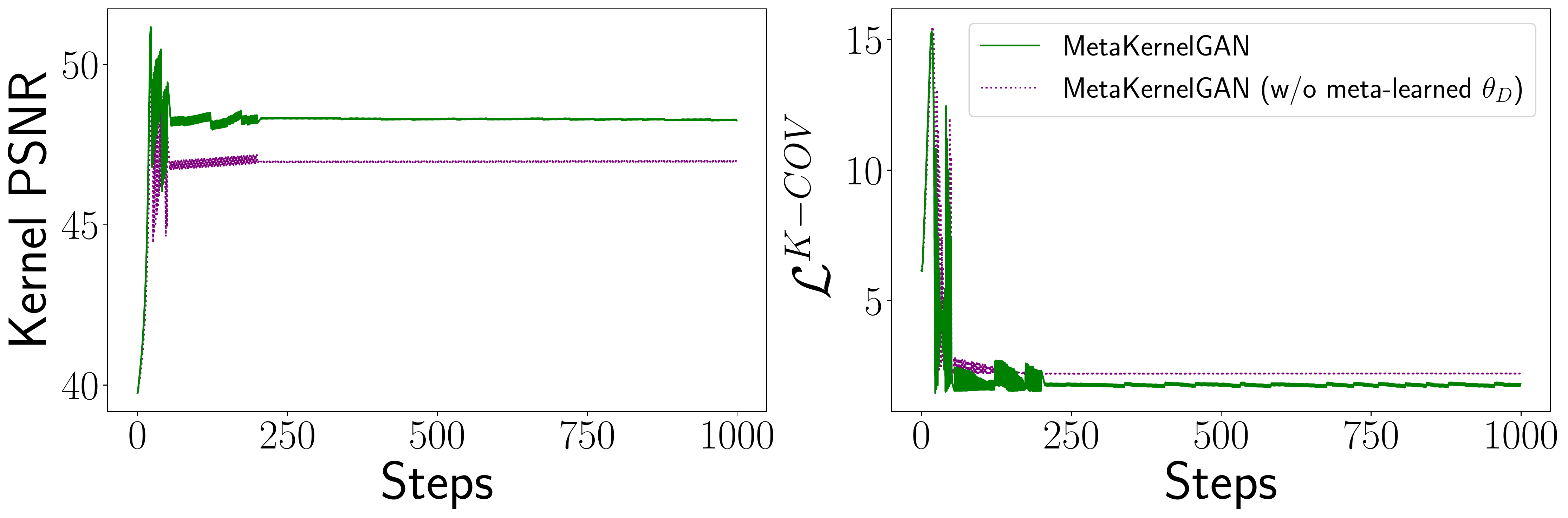}
    \hfill
    \includegraphics[width=0.47\textwidth]{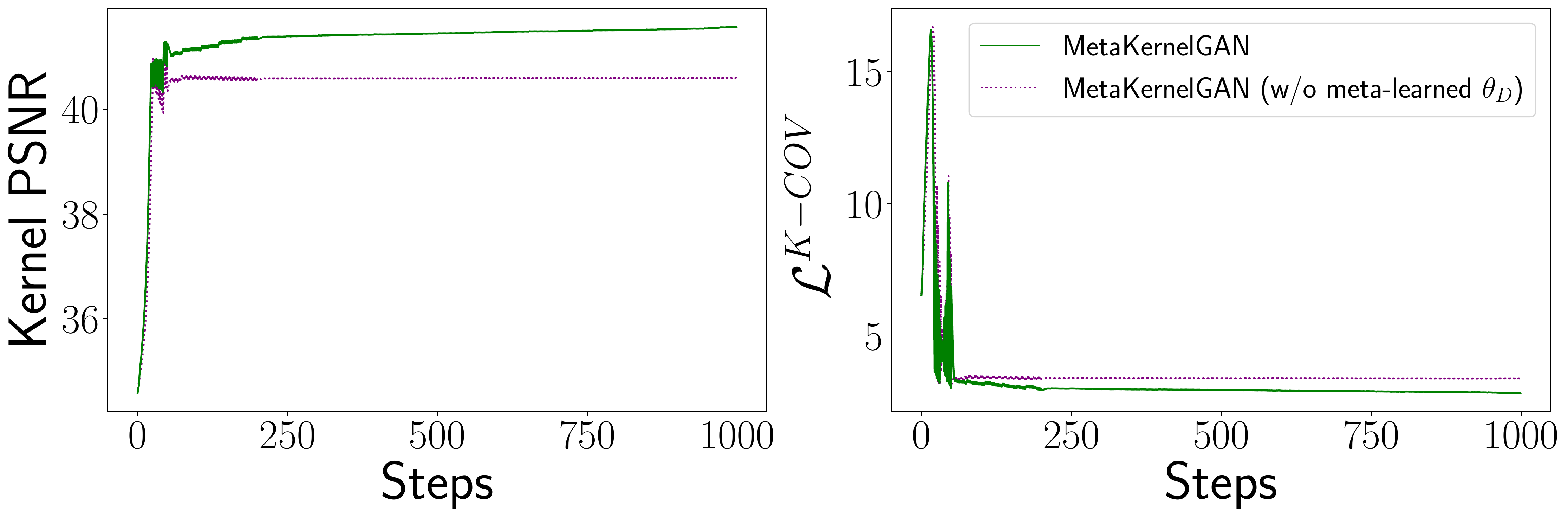}
    \hfill
    \includegraphics[width=0.47\textwidth]{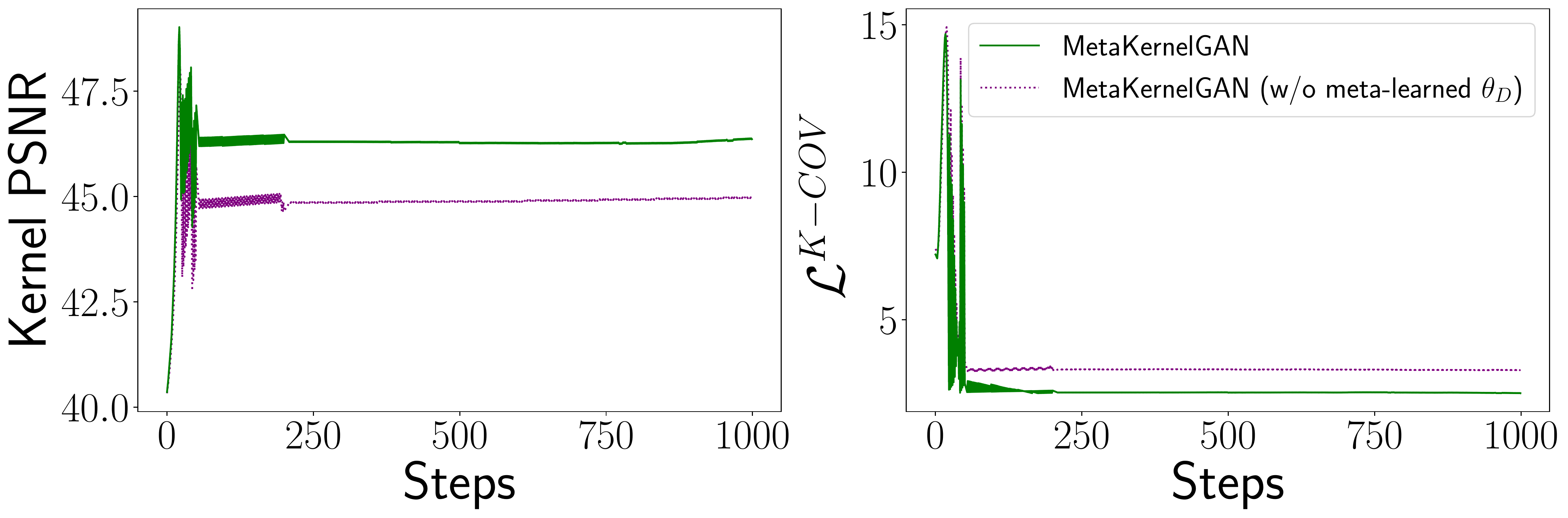}
    \hfill
    \caption{Intermediate kernel results for every adaptation iteration ($\times2$ upsampling for \textit{0802.png}, \textit{0833.png}, \textit{0838.png}, \textit{0857.png}, and \textit{0853.png} in DIV2K).}
    \label{fig:ablation}
\end{figure}

\noindent
\textbf{Meta-learning the Generator.}
Fig.~\ref{fig:ablation} shows the case where we only meta-learn the generator and not the discriminator. As the discriminator is trained from scratch for each image during evaluation, the meta-trained generator dominates the training, leading to inadequate generator feedback from the discriminator and in turn to inferior kernel accuracy. 

\noindent
\textbf{Number of Iteration Steps.}~We meta-trained MetaKernelGAN with $N_{\text{adapt}}$$=$${10,25,50}$ and showed the Kernel PSNR and $\mathcal{L}^{\text{K{\text -}COV}}$ for $\times$2 upsampling in Table.~\ref{tab:ablation_table}. We observe that the kernel has not converged when $N_{\text{adapt}}$$=$$10$, resulting in worse kernel performance. $N_{\text{adapt}}$$=$$50$ also leads to worse performance, possibly attributed to the first-order approximation of FOMAML which is not suited for long inner-loop trajectories. 

\begin{table}[t]
\renewcommand\thetable{R1}
\centering
\resizebox{0.48\textwidth}{!}{%
\begin{tabular}{l|l|l|l|l|l} \toprule
\multicolumn{1}{c}{\textbf{Method}} & \multicolumn{1}{c}{\textbf{$N_{\text{adapt}}$}} & \multicolumn{1}{c}{\textbf{Set14}} & \multicolumn{1}{c}{\textbf{B100}} & \multicolumn{1}{c}{\textbf{Urban100}} & \multicolumn{1}{c}{\textbf{DIV2K}} \\ \midrule
MetaKernelGAN & 10 & 44.72/2.56 & 43.42/2.99 & 45.99/2.39 & 46.88/2.31 \\ 
MetaKernelGAN & 25 & \textbf{46.23}/\textbf{2.44} & \textbf{45.94}/\textbf{2.38} & \textbf{47.37}/\textbf{2.16} & \textbf{48.00}/\textbf{2.08} \\
MetaKernelGAN & 50 & 45.49/2.49 & 45.41/2.65 & 46.47/2.42 & 46.81/2.37 \\ \hline
\end{tabular}%
}
\caption{Average Kernel PSNR/$\mathcal{L}^{\text{K{\text -}COV}}$ on SR benchmarks across five runs for different $N_{\text{adapt}}$ for $\times$2 upsampling.}

\label{tab:ablation_table}
\end{table}

\section{MetaKernelGAN Adaptability to Images with Limited Internal Information.}
LR images are severely limited in internal information when their resolution is small, resulting in the fallback on the learned kernel. This is often the case in \textit{i)}~$\times$4 upsampling, where the LR images are tiny, and \textit{ii)}~especially in smaller resolution datasets. To quantify this, we compare the estimated discretized covariance matrix of the adapted kernel after 200 steps, $\hat{\Sigma}^{\text{Est200}}$, to that of the initial learned kernel, $\hat{\Sigma}^{\text{Est0}}$, and that of the ground-truth kernel, $\hat{\Sigma}^{\text{GT}}$.
Specifically, we compute {\small$\mathcal{L}^{T}=\max(\mathcal{L}^{\text{K{\text -}COV}}(\hat{\Sigma}^{\text{Est200}}, \hat{\Sigma}^{\text{GT}}) - \mathcal{L}^{\text{K{\text -}COV}}(\hat{\Sigma}^{\text{Est200}},\hat{\Sigma}^{\text{Est0}}), 0)$}, where $\mathcal{L}^{\text{K{\text -}COV}}(a,b)=\sum^N_{x,y} \big| \hat{\Sigma}^{\text{a}}_{x,y} - \hat{\Sigma}^{\text{b}}_{x,y} \big|$, in three equal-sized datasets B100, Urban100, and DIV2K, representing small, medium, and large image resolutions, respectively. By definition, the higher $\mathcal{L}^{T}$ is, the closer the adapted kernel is to the initial kernel relative to its distance from the GT kernel and the higher the fallback rate. 
For $\times$2 upsampling, the mean $\mathcal{L}^{T}$ is 0.0 for all three datasets, indicating no fallback. For $\times$4, the mean $\mathcal{L}^{T}$ for B100, Urban100, and DIV2K is 4.98, 3.78, and 3.25, respectively, indicating that lower-resolution images lead to more frequent fallback.


\section{Additional Qualitative Results}
\begin{figure*}[b!]
    \centering
    \includegraphics[clip,width=0.8\textwidth]{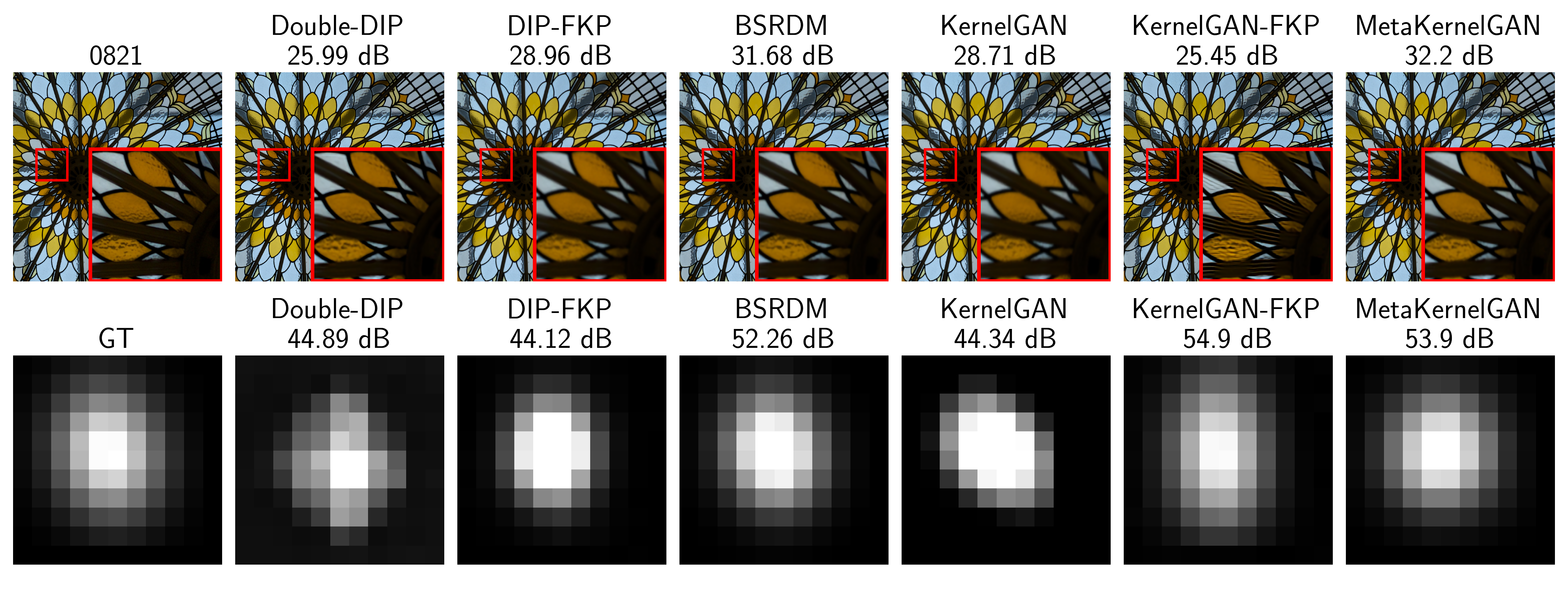}
    \hfill
    \includegraphics[ clip,width=0.8\textwidth]{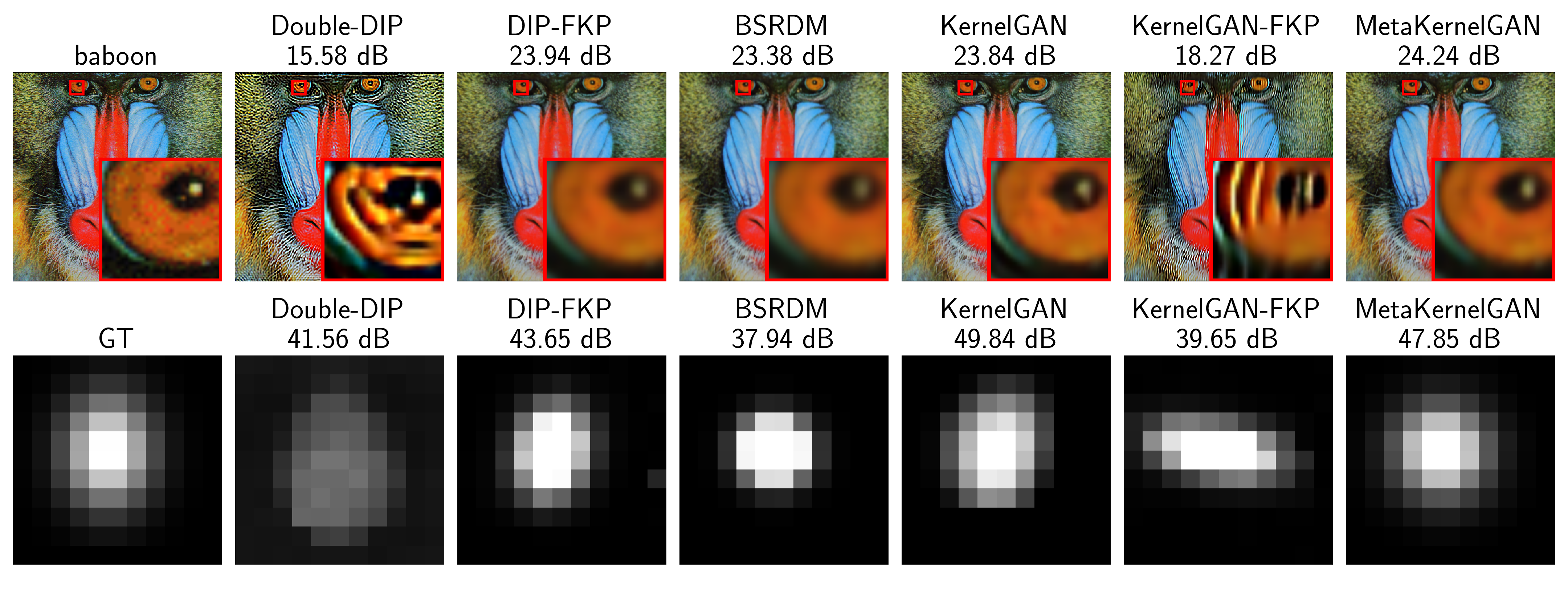}
    \hfill
    \includegraphics[clip,width=0.8\textwidth]{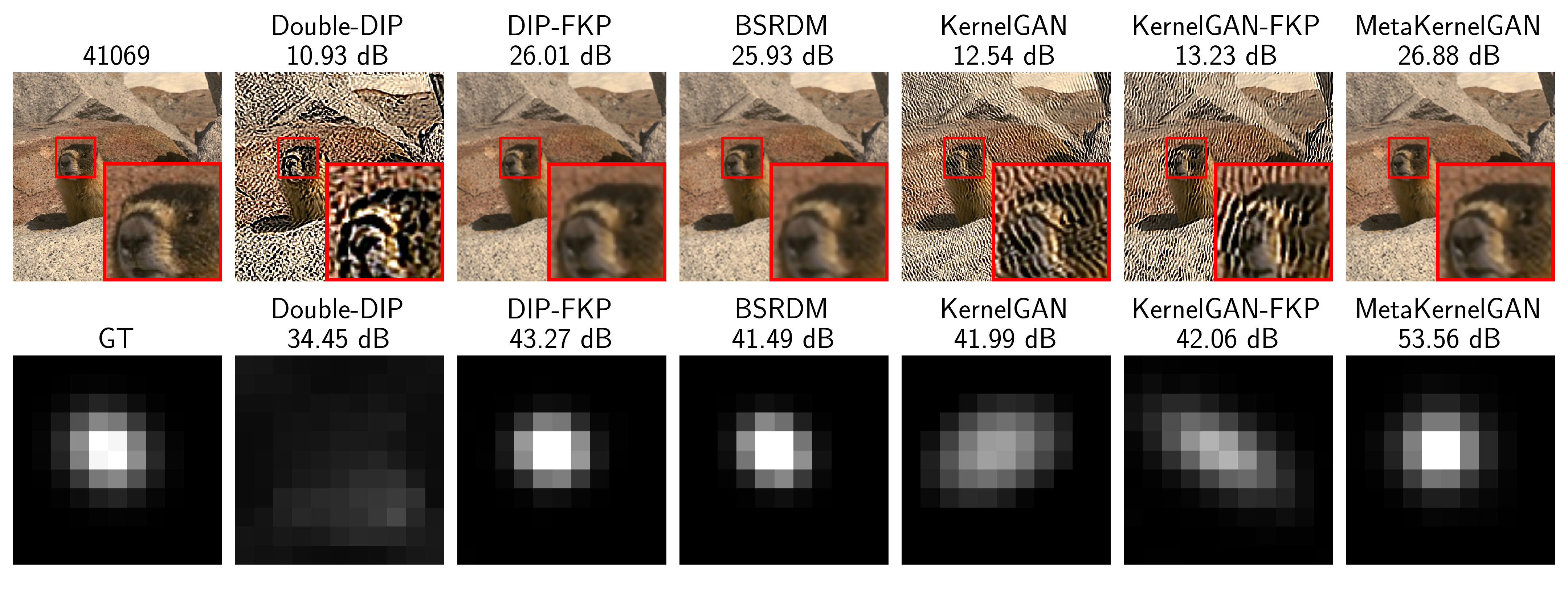}
    \hfill
    \caption{Comparison of estimated kernel, along with its image $\times2$ upsampled using USRNet~\cite{USRNet}, among explicit kernel estimation methods across different benchmark datasets. Zoom in for best results.}
    \label{fig_sm:img_w_kerx2}
\end{figure*}

\begin{figure}[b!]
    \centering
    \includegraphics[ clip,width=0.47\textwidth]{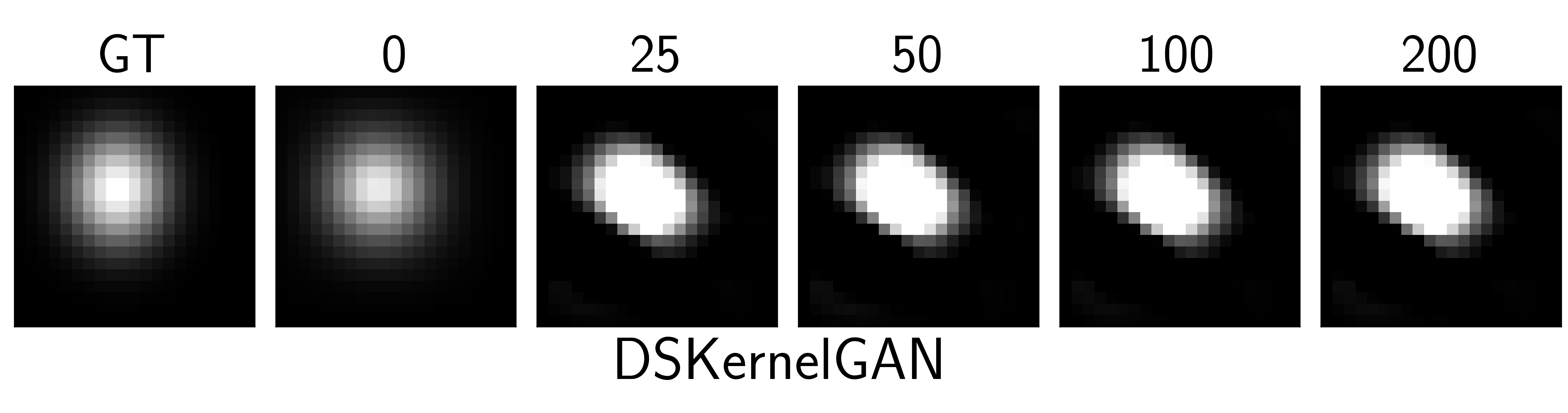}
    \includegraphics[clip,width=0.47\textwidth]{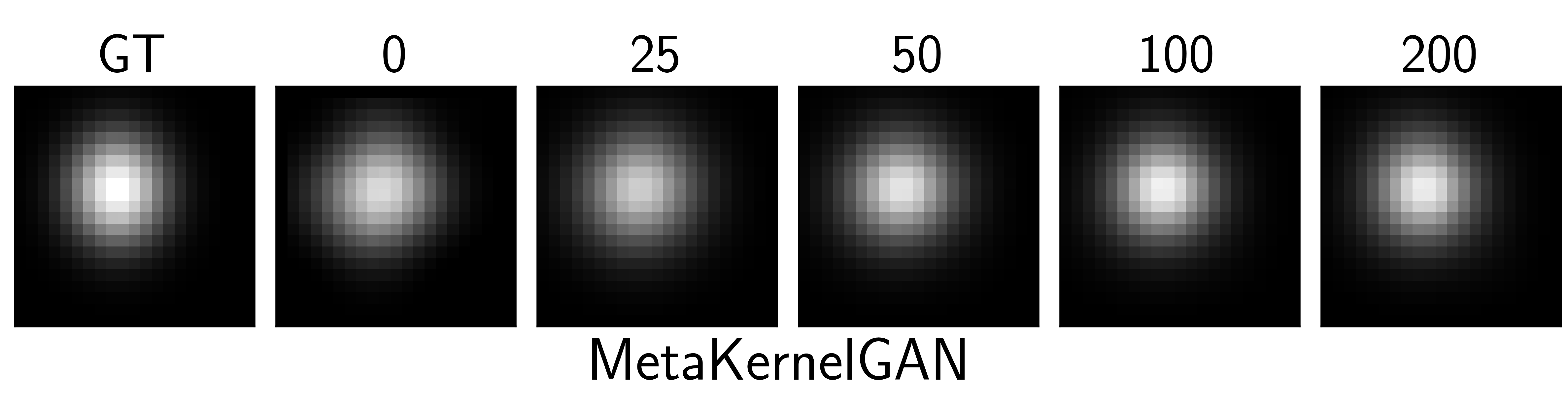}
    \hrule
    \includegraphics[ clip,width=0.47\textwidth]{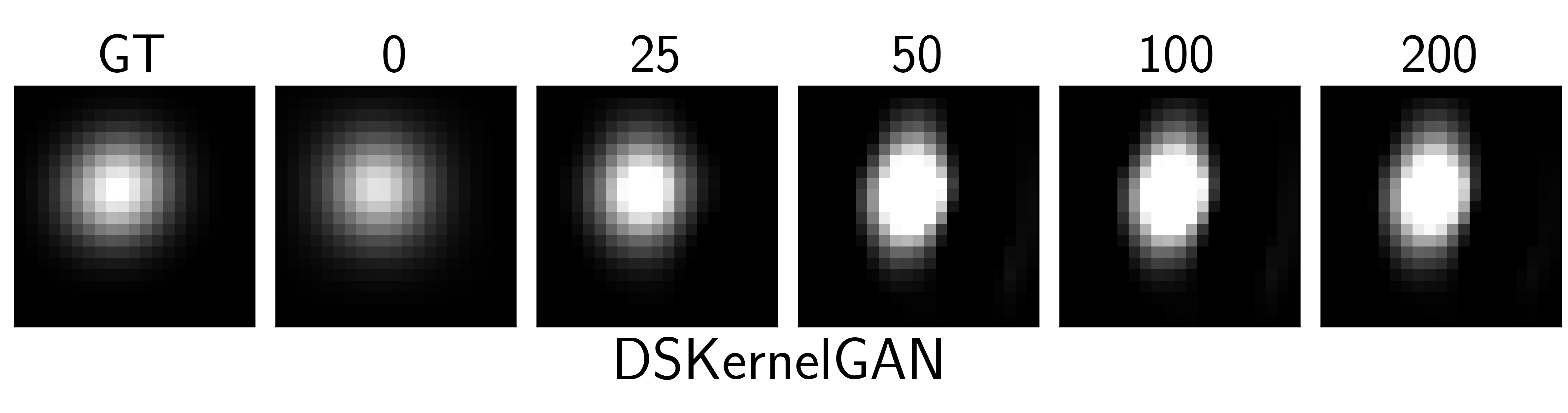}
    \includegraphics[clip,width=0.47\textwidth]{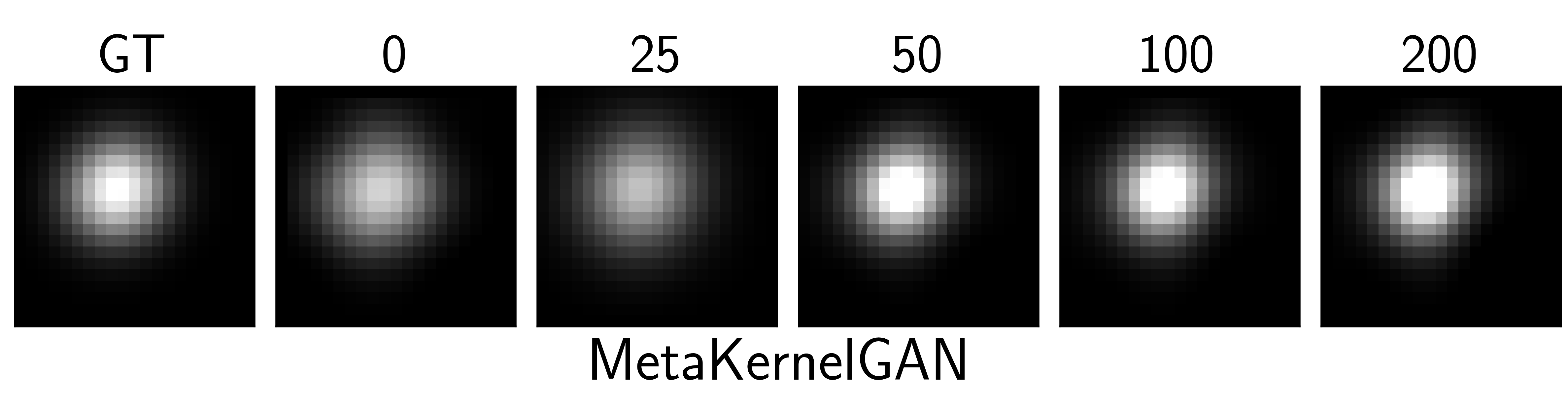}
    \hrule
    \includegraphics[ clip,width=0.47\textwidth]{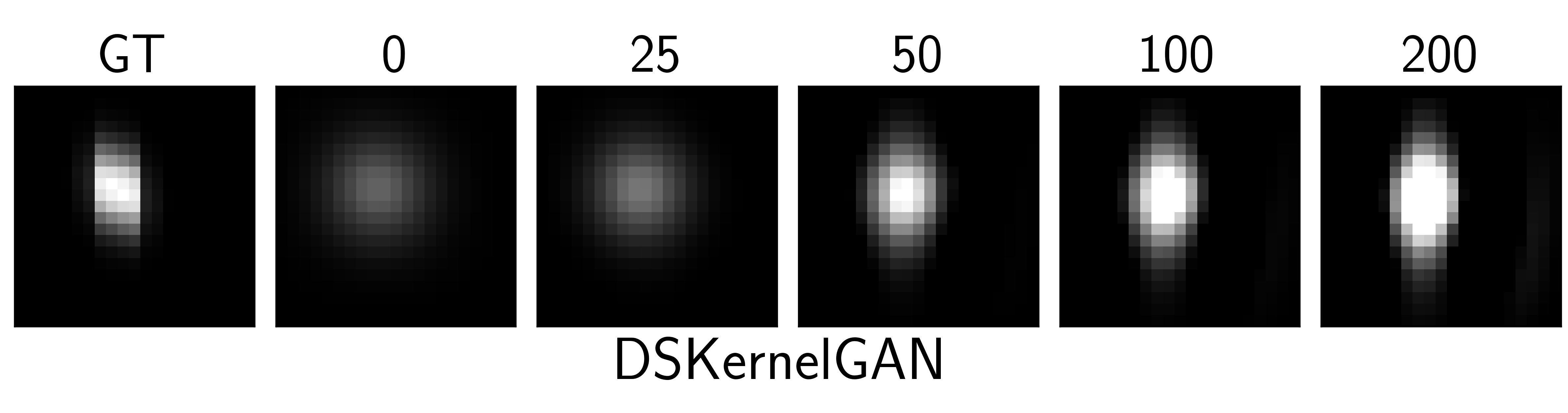}
    \includegraphics[clip,width=0.47\textwidth]{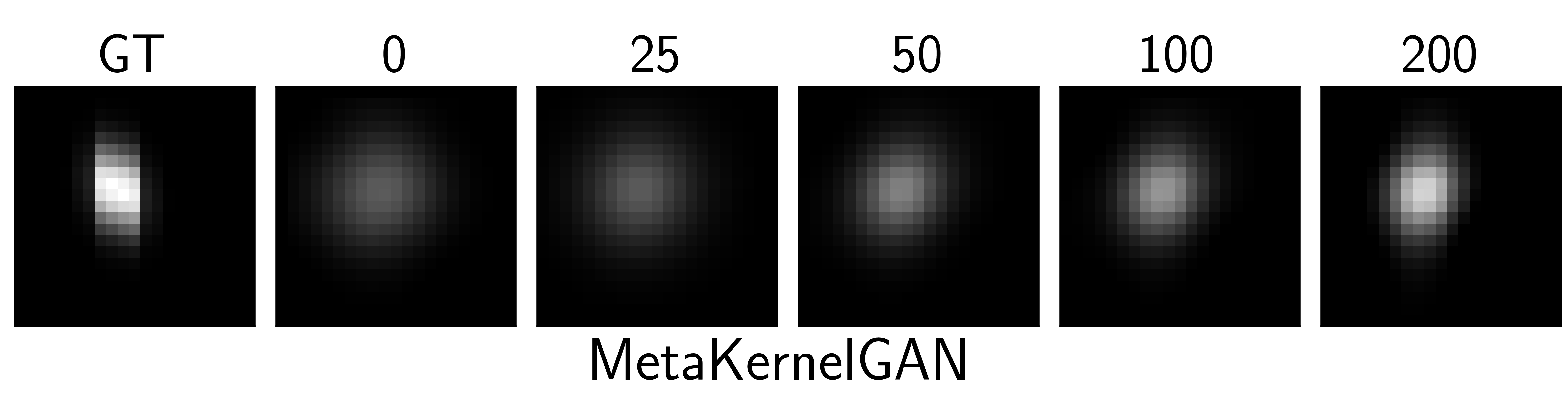}
    \hrule
    \includegraphics[clip,width=0.47\textwidth]{latex/dskg_mkg_figs/0824.png_x4_kernel_adaptation_DSKernelGAN.png}
    \includegraphics[clip,width=0.47\textwidth]{latex/dskg_mkg_figs/0824.png_x4_kernel_adaptation_MetaKernelGAN.png}
    \caption{Kernel after 25, 50, 100, 200 adaptation steps for $\times$4 upsampling on \textit{monarch} and \textit{ppt3} of Set14, and \textit{0879} and \textit{0824} of DIV2K (top to bottom).}
    \label{fig_sm:adaptation_stepx4}
\end{figure}

\begin{figure*}[b!]
    \centering
    
    \includegraphics[ clip,width=0.6\textwidth]{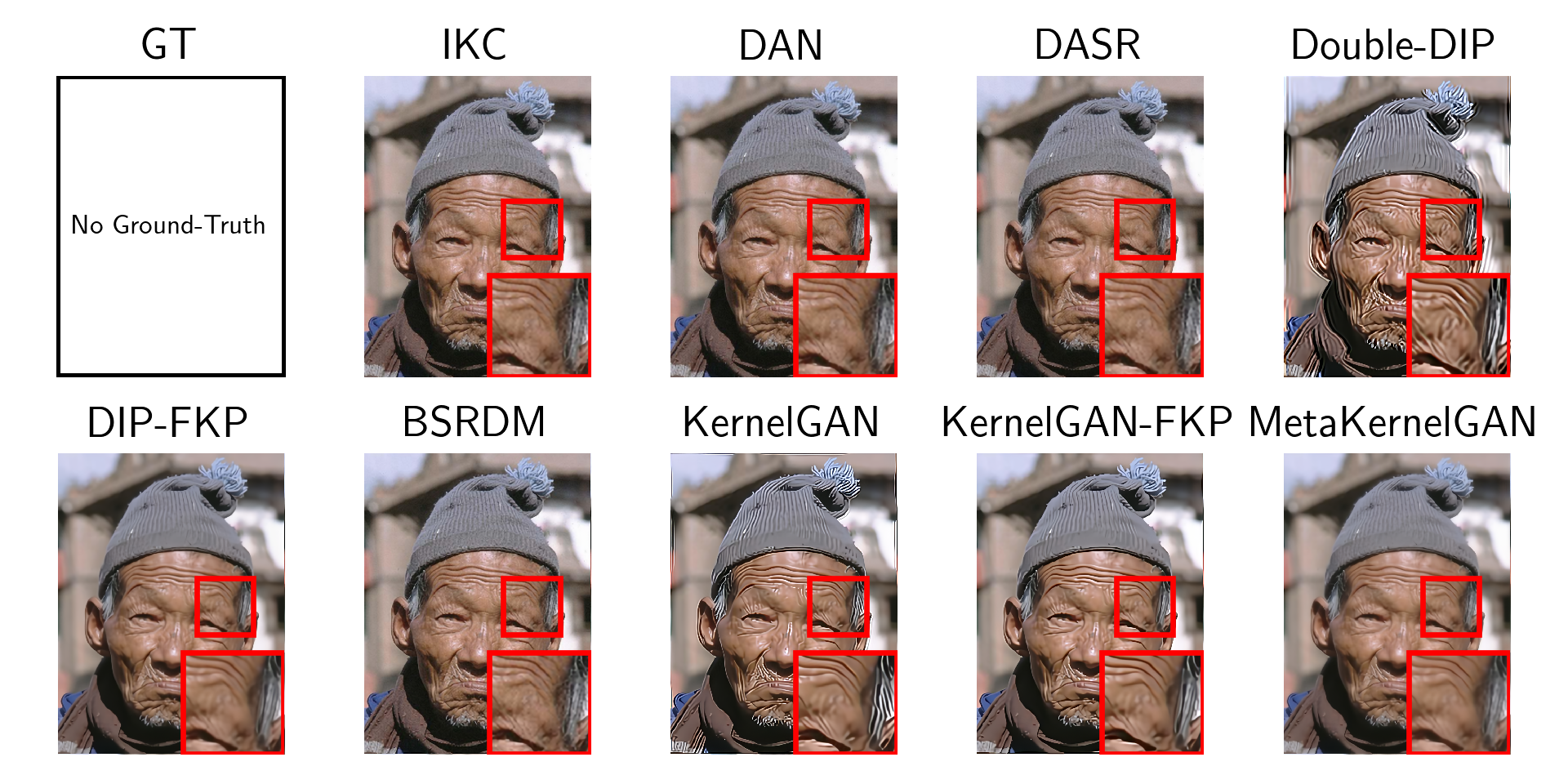}
    \hfill
    \includegraphics[clip,width=0.8\textwidth]{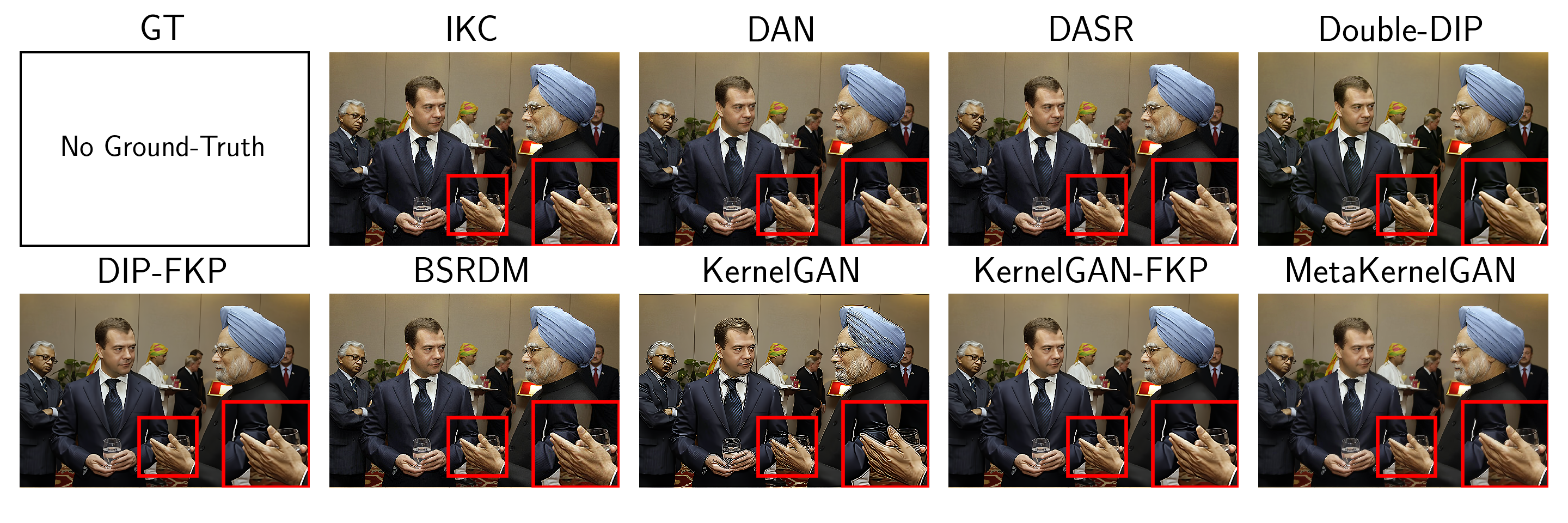}
    \hfill
    \includegraphics[clip,width=0.8\textwidth]{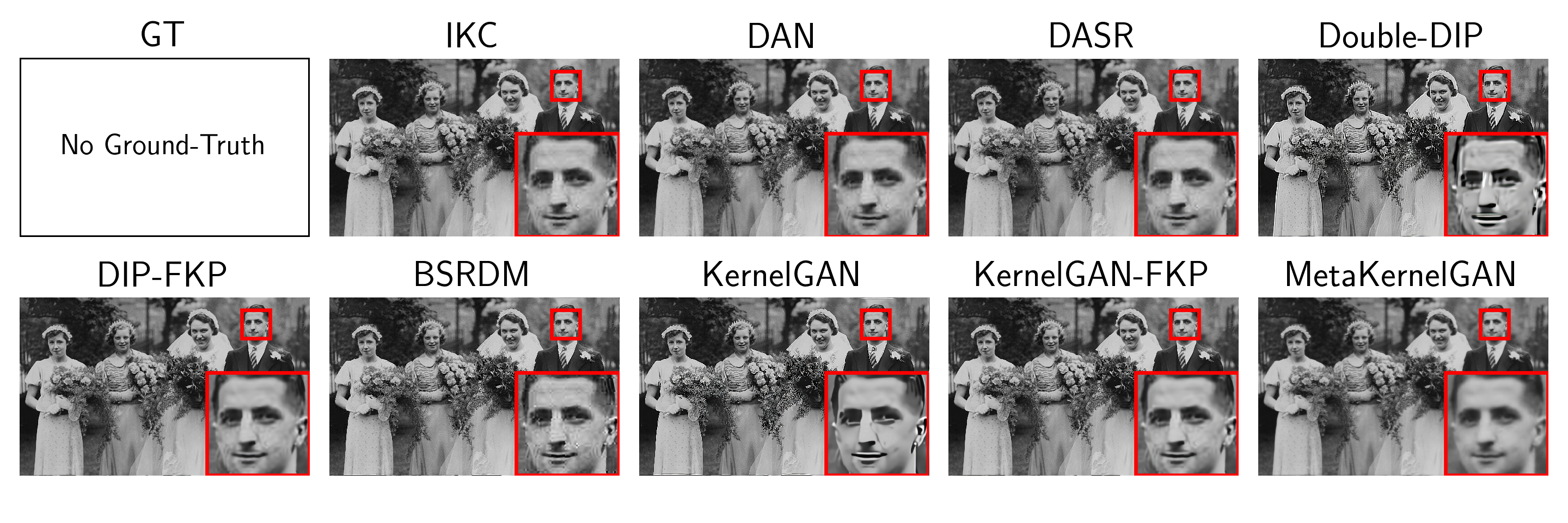}
    \hfill
    \includegraphics[clip,width=0.8\textwidth]{latex/figs_png/n_real_world_pexels-suzy-hazelwood-6707072_x4.png}
    \caption{Real-world visual quality comparison on $\times4$ upsampling among models. Zoom in for best results. No ground truth is available.}
    \label{fig_sm:real_world}
\end{figure*}

Fig.~\ref{fig_sm:img_w_kerx2} \& Fig.~\ref{fig_sm:img_w_kerx4} show the comparisons of both the kernel and image among the explicit kernel estimation methods on our benchmark evaluation datasets for $\times2$ and $\times4$ upsampling respectively.
Fig.~\ref{fig_sm:adaptation_stepx4} show more examples highighting that DSKernelGAN is more susceptible to adapt to faulty kernels than MetaKernelGAN as the former doesn't learn from the adaptation process.  
Lastly, we show more real-world results among both implicit and explicit degradation methods from images downloaded from the Internet in Fig.~\ref{fig_sm:real_world}. 


\begin{figure*}[b!]
    \centering
    
    \includegraphics[ clip,width=0.8\textwidth]{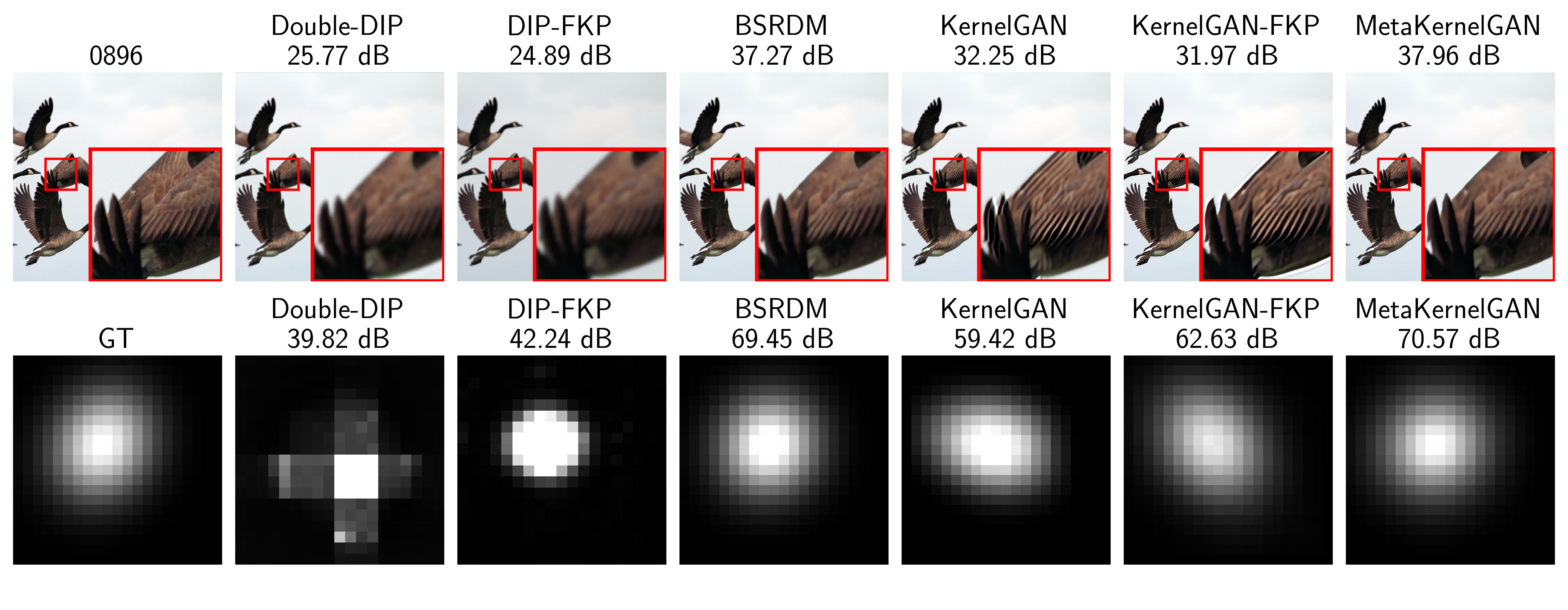}
    \hfill
    \includegraphics[clip,width=0.8\textwidth]{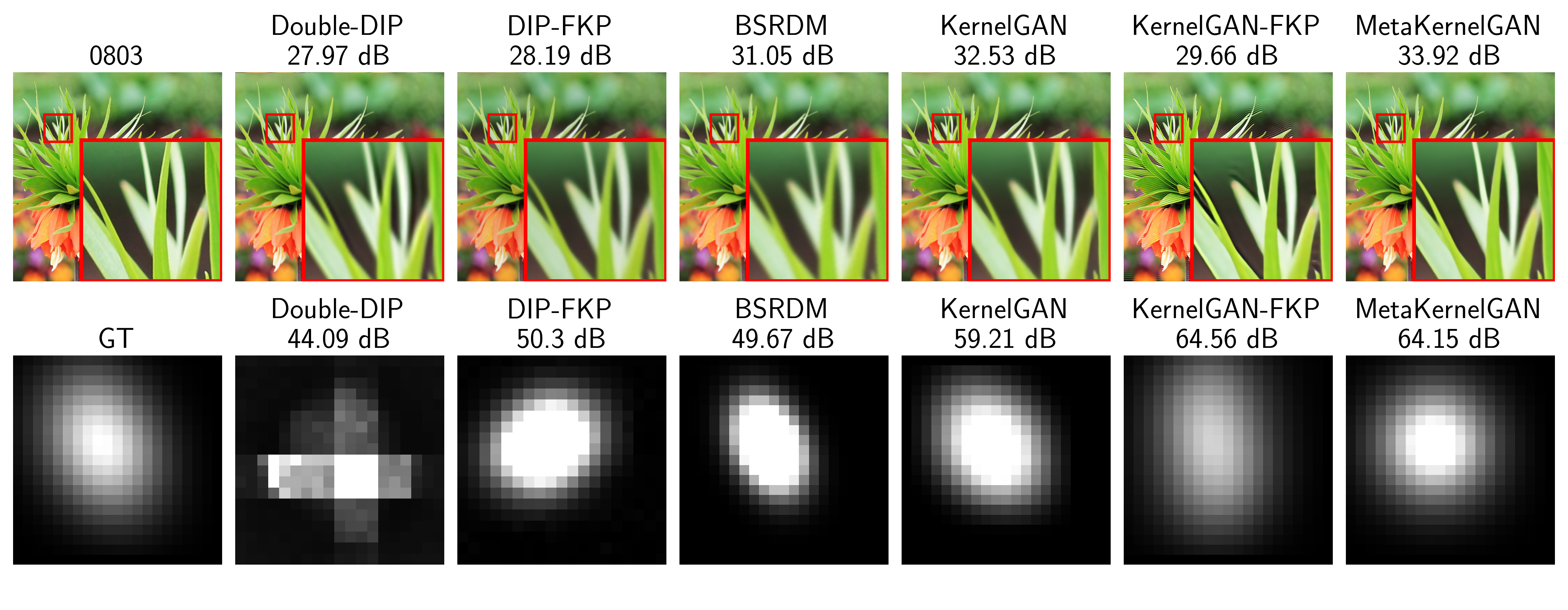}
    \hfill
    \includegraphics[clip,width=0.8\textwidth]{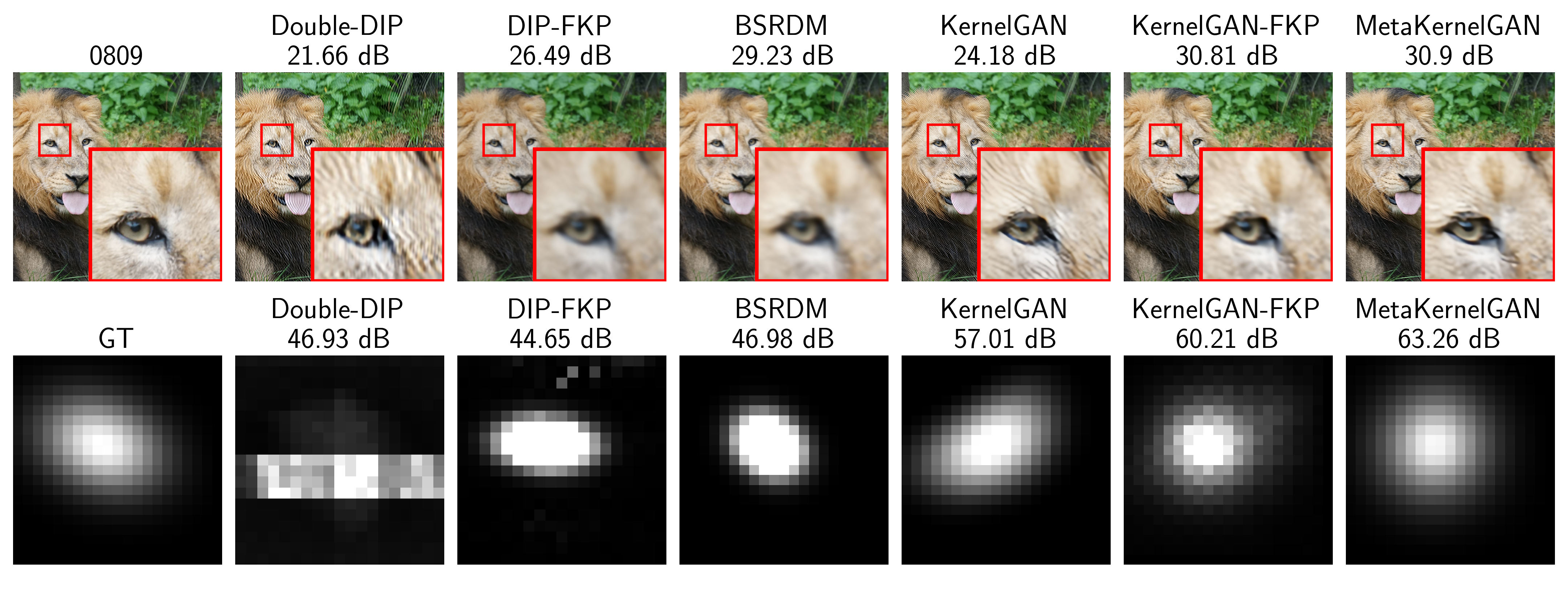}
    \caption{Comparison of estimated kernel, along with its image $\times4$ upsampled using USRNet~\cite{USRNet}, among explicit kernel estimation methods across different benchmark datasets. Zoom in for best results. Part 1 of 2.}
\end{figure*}
\begin{figure*}[ht]\ContinuedFloat
    \centering
    \includegraphics[clip,width=0.8\textwidth]{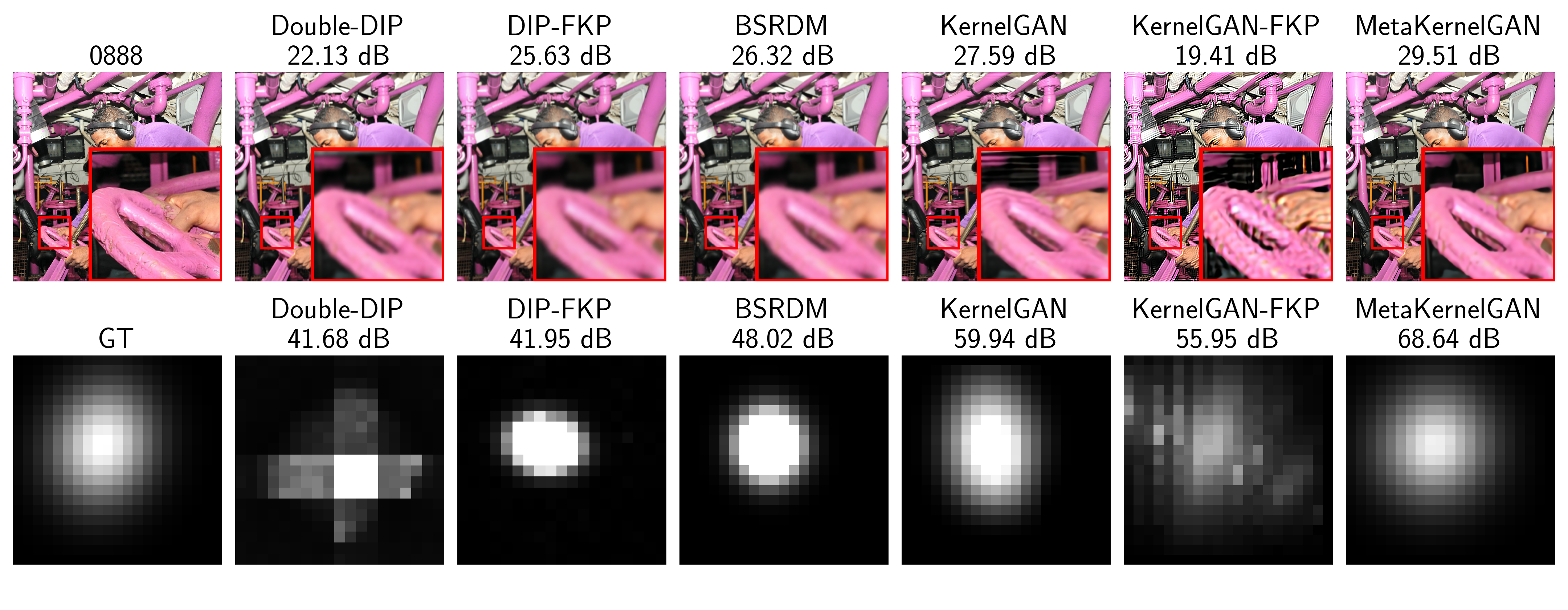}
    \hfill
    \includegraphics[clip,width=0.8\textwidth]{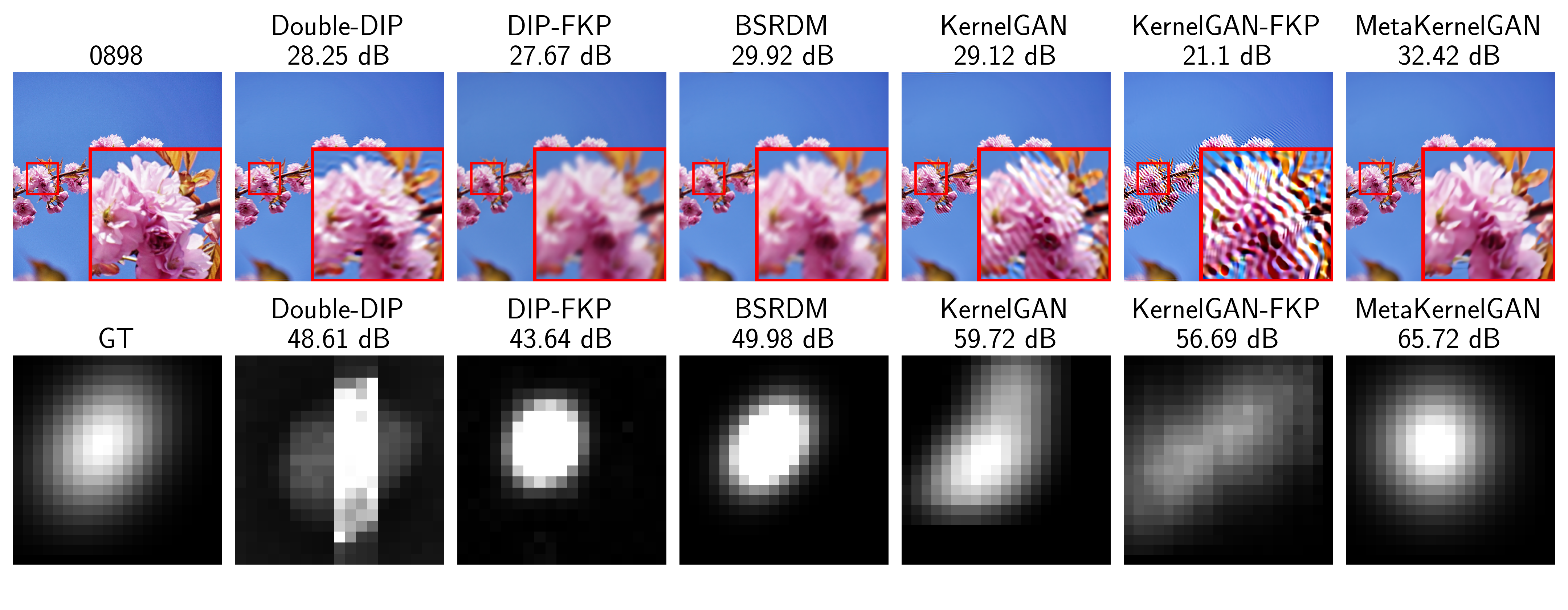}
    \hfill
    \caption{Comparison of estimated kernel, along with its image $\times4$ upsampled using USRNet~\cite{USRNet}, among explicit kernel estimation methods across different benchmark datasets. Zoom in for best results. Part 2 of 2.}
    \label{fig_sm:img_w_kerx4}
\end{figure*}

\newpage
\begin{figure}[b!]
    \centering
    \includegraphics[ clip,width=0.47\textwidth]{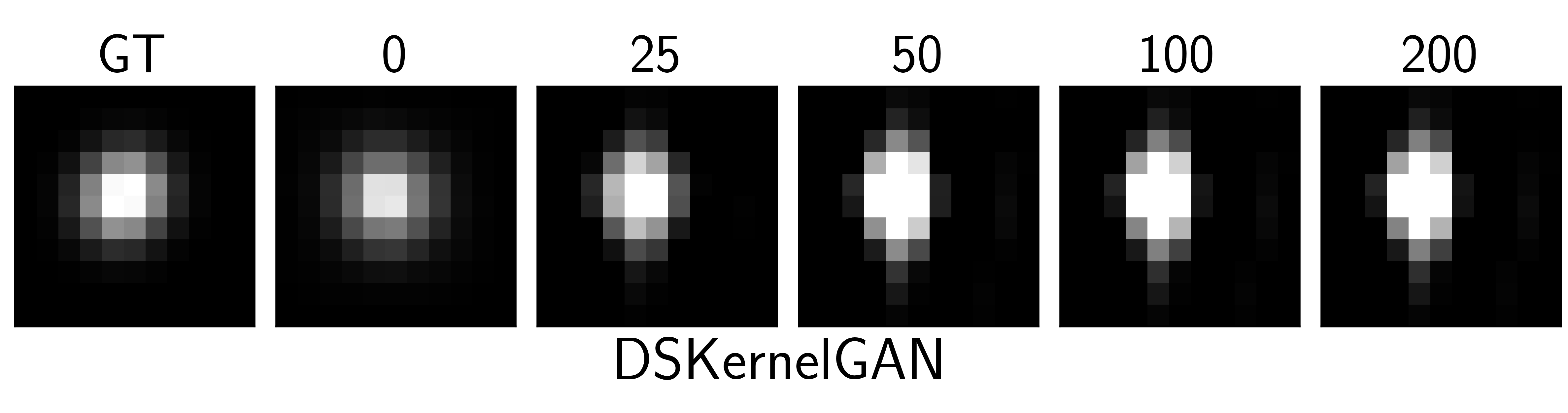}
    \includegraphics[clip,width=0.47\textwidth]{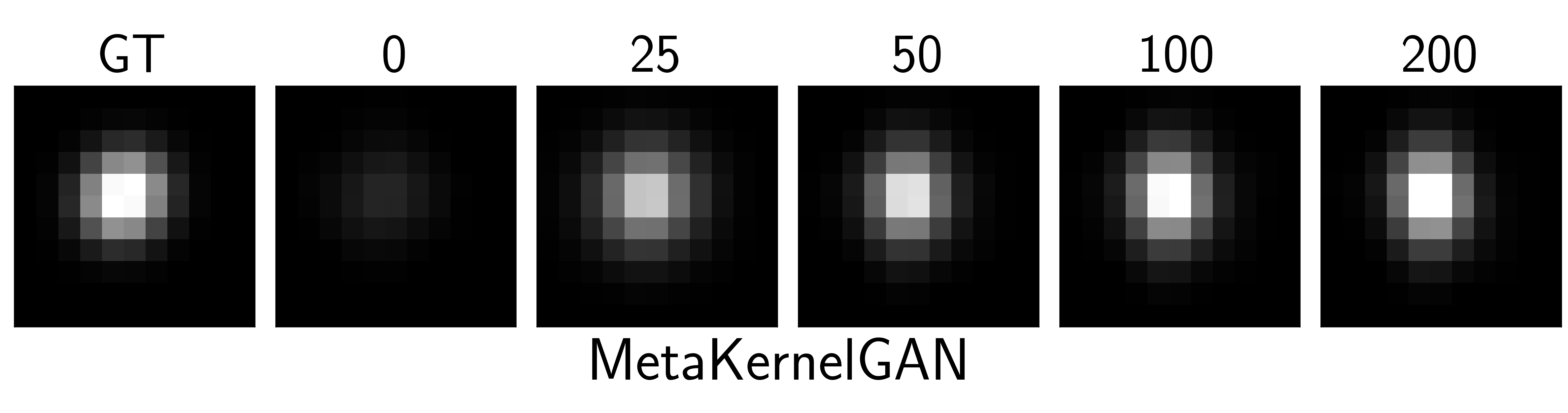}
    \hrule
    \includegraphics[ clip,width=0.47\textwidth]{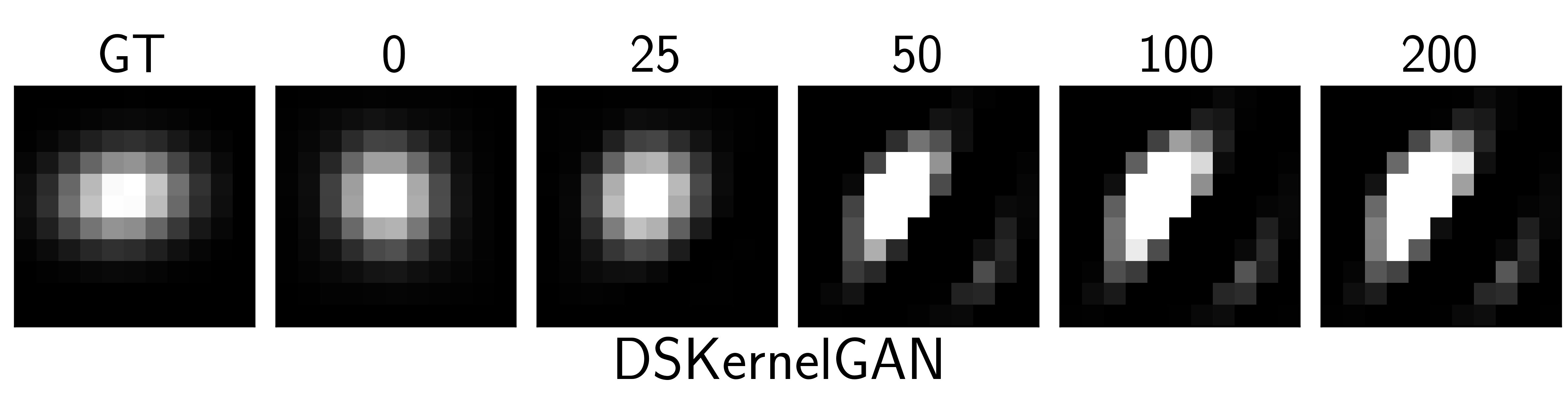}
    \includegraphics[clip,width=0.47\textwidth]{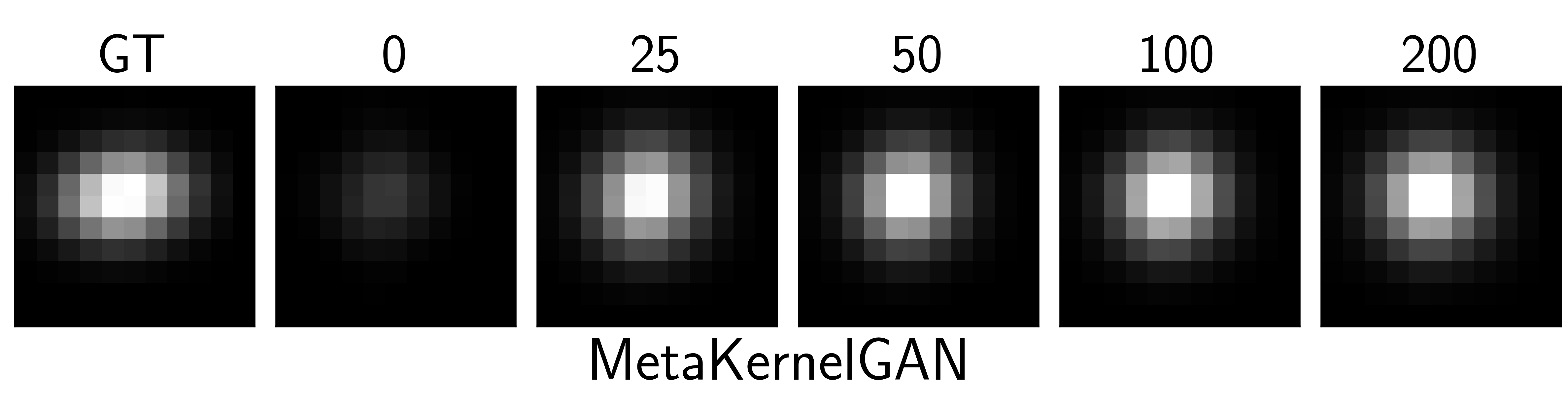}
    \hrule
    \includegraphics[ clip,width=0.47\textwidth]{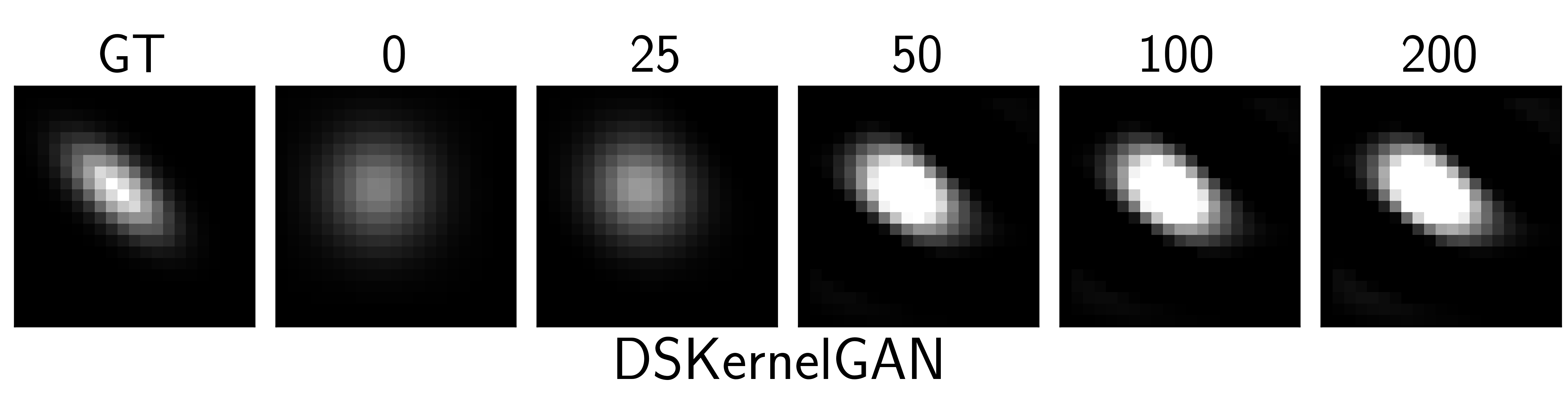}
    \includegraphics[clip,width=0.47\textwidth]{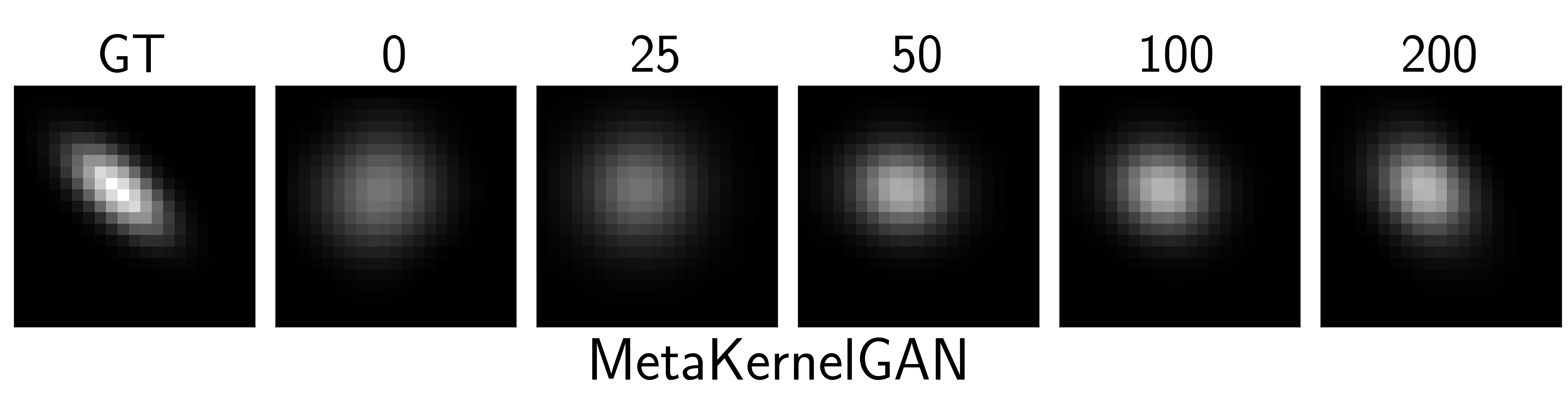}
    \hrule
    \includegraphics[clip,width=0.47\textwidth]{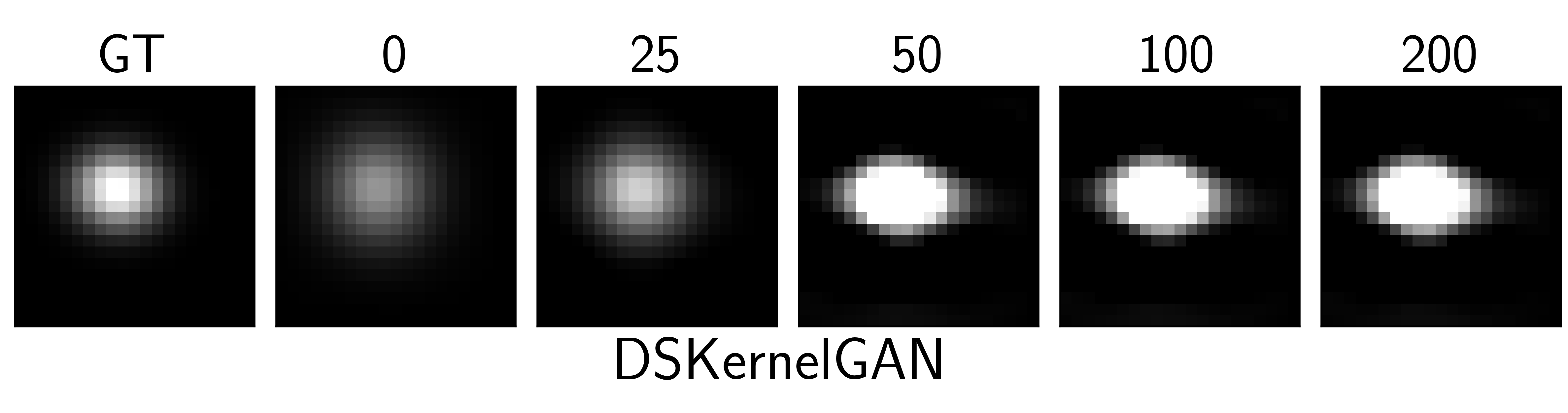}
    \includegraphics[clip,width=0.47\textwidth]{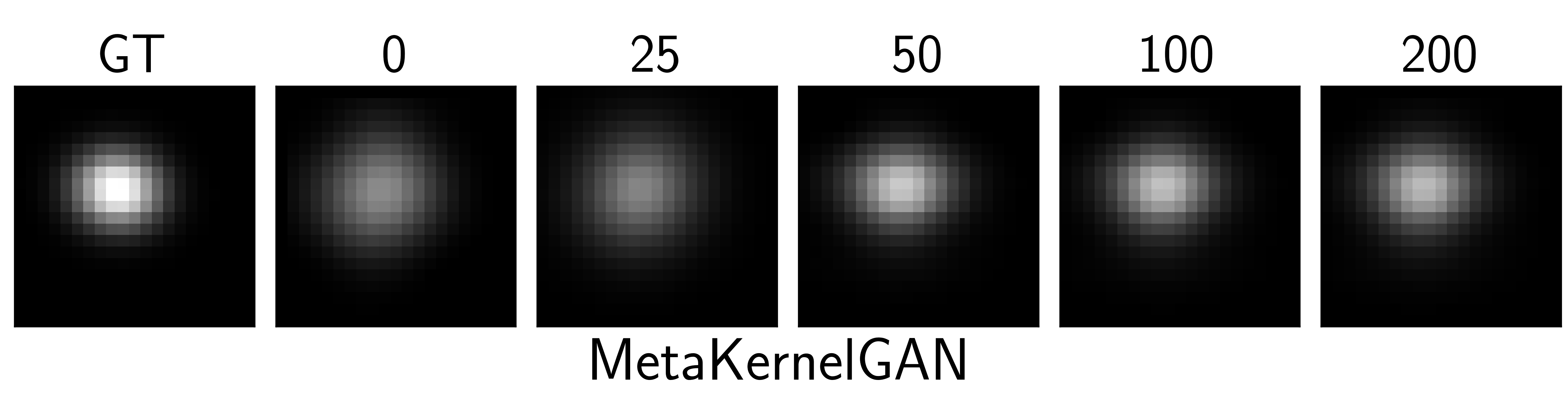}
    \caption{Kernel after 25, 50, 100, 200 adaptation steps for $\times$2 upsampling on \textit{86000} of B100, \textit{img036} of Urban100, \textit{0821} and \textit{0824} of DIV2K (top to bottom).}
    \label{fig_sm:adaptation_stepx2}
\end{figure}

\end{document}